\documentclass[lettersize,journal]{IEEEtran}
\usepackage{amsmath,amsfonts}
\usepackage{algorithm}
\usepackage{algorithmic}
\usepackage{array}
\usepackage[caption=false,font=normalsize,labelfont=sf,textfont=sf]{subfig}
\usepackage{textcomp}
\usepackage{stfloats}
\usepackage{url}
\usepackage{verbatim}
\usepackage{graphicx}
\usepackage{svg}
\usepackage{cite}
\usepackage{color}

\usepackage{cite}
\usepackage{xurl}
\def\BibTeX{{\rm B\kern-.05em{\sc i\kern-.025em b}\kern-.08em
    T\kern-.1667em\lower.7ex\hbox{E}\kern-.125emX}}

\DeclareMathOperator*{\argmax}{\text{arg max}}

\begin{document}

\title{A Bayesian Framework for Active Tactile Object Recognition, Pose Estimation and Shape Transfer Learning}

\author{Haodong Zheng, Andrei C. Jalba, Raymond H. Cuijpers, Wijnand A. IJsselsteijn, Sanne Schoenmakers 
\thanks{This work was partially supported by the EMDAIR project I.Touch2See funded by the Eindhoven Artificial Intelligence Systems Institute (EAISI) of the Eindhoven University of Technology (TU/e).}
\thanks{The authors are with Eindhoven University of Technology, 5612 AZ Eindhoven, The Netherlands. (email addresses: h.zheng3@tue.nl, a.c.jalba@tue.nl, r.h.cuijpers@tue.nl, w.a.ijsselsteijn@tue.nl, s.schoenmakers@tue.nl)
}}

\markboth{IEEE Transactions ON XXXXXX,~Vol.~XX, No.~XX, MONTH~YEAR}%
{Shell \MakeLowercase{\textit{et al.}}: A Sample Article Using IEEEtran.cls for IEEE Journals}

\maketitle

\begin{abstract}
In this paper, we combine a customized particle filter (PF) and a Gaussian process implicit surface (GPIS) to address active tactile object recognition, pose estimation, and shape transfer learning in a unified Bayesian framework. The PF maintains a joint posterior over object class and pose with a progressive sampling strategy to improve sampling efficiency and keeps inference tractable during active tactile exploration. Novel objects are identified through MAP model evidence, after which GPIS reconstructs their shape using the PF maximum-a-posteriori (MAP) estimate as prior, enabling transfer of geometric knowledge from known objects. An exploration strategy based on global shape estimates guides data acquisition and automatically terminates when sufficient data coverage on the estimated surface is achieved. Simulation experiments demonstrate effective recognition of known objects, reconstruction of novel shapes, and reliable re-identification after learning.
\end{abstract}


\section{Introduction}

\IEEEPARstart{T}{actile} sensing is essential in robotic perception, serving both as a complementary sense to vision and as an independent modality. In unstructured environments, visual information can be limited or unreliable due to severe occlusion, poor visibility conditions, or unmodeled interaction effects. Under situations where vision perception is unreliable or unavailable, tactile-based perception methods are capable of functioning independently of visual input and provide a robust solution for robotic perception. Humans can perceive the object's identity, pose and shape by touch only \cite{klatzky1985identifying}. Similar capabilities would also be desirable for robotic systems, as understanding an object's shape and pose is essential for accurate positioning and effective interaction with the object, making it valuable for a wide range of applications.

However, tactile observations of robots are inherently local and sparse: they do not provide sufficient information to disambiguate an object's class, pose, and shape in a single touch. This ambiguity makes active exploration necessary and calls for inference methods that explicitly keep track of uncertainty and reason over sequential data from active exploration.
 


In many real-world scenarios, the robot may encounter both known and novel objects, two cases that are typically handled differently. For known objects, the object class and pose should be estimated, whereas novel objects require shape exploration and learning. Existing tactile perception systems typically address these tasks separately: Either address object recognition and pose estimation assuming the object belongs to a known prior object set or reconstruct the shape of object with object-agnostic priors without explicit mechanism for novelty detection or knowledge transfer. This separation limits the robot's ability to reason about novelty of the object and draw connection to the existing prior knowledge for efficient learning.

In this study, we propose a unified Bayesian framework that jointly reasons about the object class, pose, and shape through active exploration from initial contact to conclusion. 
Our framework maintains a joint belief distribution over object class and 6-DOF pose using a customized particle filter (PF) designed to remain tractable. By tracking Bayesian model evidence of the particles, the system can detect novel objects when all particles yield low model evidence. When novelty is detected, the current maximum a posteriori (MAP) estimation from the PF is used to initialize Gaussian Process Implicit Surface (GPIS) \cite{williamsGaussianProcessImplicit2006} for shape reconstruction, enabling shape estimation with uncertainty and allowing prior knowledge from known shapes to transfer to the novel-object reconstruction. The evolving global shape estimate (from either the MAP estimation or GPIS) guides active exploration, and a directed Hausdorff distance (DHD) criterion determines when exploration should terminate automatically.

The main contributions of this paper are:
\begin{itemize}
\item A tractable customized PF for joint inference on object class and 6-DOF pose by progressive sampling new particles based on point-pair features\cite{drostModelGloballyMatch2010} of the tactile contact points;
\item A unified Bayesian framework that transfers knowledge from known shapes to learn novel shapes using PF-MAP-initialized GPIS and simultaneously estimates the class and pose for known objects and novel objects;
\item An active tactile exploration procedure guided by global shape estimates, with automatic termination using a DHD-based criterion.
\end{itemize}

\section{Related Work}

The proposed framework relates to belief-space inference for tactile perception, probabilistic surface modeling, and active decision-making under uncertainty. A common challenge across these domains is maintaining tractable inference in high-dimensional and partially observable settings while enabling closed-loop exploration.

\subsection{Bayesian Tactile Recognition and Pose Estimation}

Early tactile recognition researches distinguished objects using material properties such as stiffness, texture, and thermal conductivity \cite{xuTactileIdentificationObjects2013,kaboliTactileBasedFrameworkActive2017,kaboliTactilebasedActiveObject2019}. In contrast, our work focuses on geometric information, which directly supports pose estimation and shape reasoning. 
Martinez-Hernandez et al. \cite{8012465} proposed an active Bayesian framework for tactile object recognition, demonstrating closed-loop uncertainty reduction through exploratory actions. Their formulation focuses on recognition rather than joint inference over object identity and pose.

Bayesian methods for touch-based pose estimation typically assume a known object model. The manifold particle filter \cite{kovalPoseEstimationPlanar2015} takes advantage of contact constraints to improve sampling efficiency. The Scaling Series algorithm \cite{petrovskayaGlobalLocalizationObjects2011} combines Monte Carlo sampling with annealing to refine 6-DOF pose estimates. Vezzani et al. \cite{vezzaniMemoryUnscentedParticle2017} proposed a memory unscented Kalman filter (MUKF) to represent pose hypotheses as a set of Gaussian distributions. While effective, these approaches address pose estimation for known objects and do not explicitly represent uncertainty over object identity.

Vezzani et al. \cite{vezzaniNovelBayesianFiltering2016} extended localization to recognition by evaluating pose estimation across multiple object models and selecting the best match. In this formulation, a Kalman update step is required for each particle, which is computationally costly. 
Inspired by manifold filtering \cite{kovalPoseEstimationPlanar2015} and point-pair features \cite{drostModelGloballyMatch2010}, we employ progressive sampling to maintain tractable inference as new observations are obtained. Unlike \cite{vezzaniNovelBayesianFiltering2016}, our approach keeps the number of particles small without requiring Kalman update steps, and it is able to revisit states ruled out by early partial observations, which is important for finding the overall best prior to reconstruct novel shapes.

\subsection{Probabilistic Shape Reconstruction}

Gaussian Processes (GPs) are widely used for tactile shape reconstruction due to their ability to model surface uncertainty. Gaussian Process Implicit Surfaces (GPIS) \cite{williamsGaussianProcessImplicit2006} have been adopted for active touch-based reconstruction \cite{meierProbabilisticApproachTactile2011,dragievGaussianProcessImplicit2011, 6630564,liDexterousGraspingShape2016,driessActiveLearningQuery2017,driessActiveMultiContactContinuous2019}. These approaches generally rely on fixed, object-agnostic priors. Martens et al. \cite{martensGeometricPriorsGaussian2017} introduced a parametric ellipsoid prior fitted to observed dense visual point cloud data, but they did not tackle sparse sequential tactile input. 

In contrast, our approach combines particle filtering and GPIS such that the evolving object class and pose posterior provides an adaptive prior for surface reconstruction throughout active tactile exploration. This enables probabilistic transfer of geometric knowledge from known objects to novel objects under a unified probabilistic framework.

\subsection{Active Exploration Under Uncertainty}

Active tactile perception is essential due to the locality of contact observations. Martinez-Hernandez et al. \cite{8012465} proposed an active tactile object recognition strategy guided by an interestingness metric. Information-theoretic criteria have been used for object recognition, selecting actions that minimize entropy or confusion \cite{schneiderObjectIdentificationTactile2009,xuTactileIdentificationObjects2013,kaboliTactileBasedFrameworkActive2017,kaboliTactilebasedActiveObject2019}. For shape reconstruction, GP-based exploration typically targets points of maximal predictive uncertainty or mutual information \cite{jamaliActivePerceptionBuilding2016,yiActiveTactileObject2016,yangLevelSetBasedGreedy2017}. Extensions in GP-based exploration also incorporate travel cost \cite{matsubaraActiveTactileExploration2017} or continuous multi-finger exploration \cite{driessActiveLearningQuery2017,driessActiveMultiContactContinuous2019}.

Unlike prior work that treats object recognition, pose estimation and novel shape reconstruction separately, our method uses the global belief (MAP or GPIS) to guide exploration and employs a directed Hausdorff distance (DHD) criterion for automatic termination, enabling unified decision-making across tasks.

\subsection{Deep-Learning-Based Approaches}

Recent deep-learning-based methods reconstruct object shapes by leveraging priors encoded in pre-trained neural networks. DeepSDF-style formulations \cite{comi2024touchsdf} and visuo-tactile models \cite{wang3DShapePerception2018,rustlerEfficientVisuoHapticObject2023a,smith3DShapeReconstruction2020,smith2021active} leverage large datasets and learned embeddings, sometimes combined with reinforcement learning for exploration. These approaches typically rely on fixed priors and do not maintain explicit probabilistic beliefs over object identity and pose.

In contrast, our approach explicitly addresses the uncertainty in object identity, pose, and shape without requiring large-scale training data, under a unified probabilistic formulation.

\begin{figure*}[!t]
\centerline{\includegraphics[width = \linewidth]{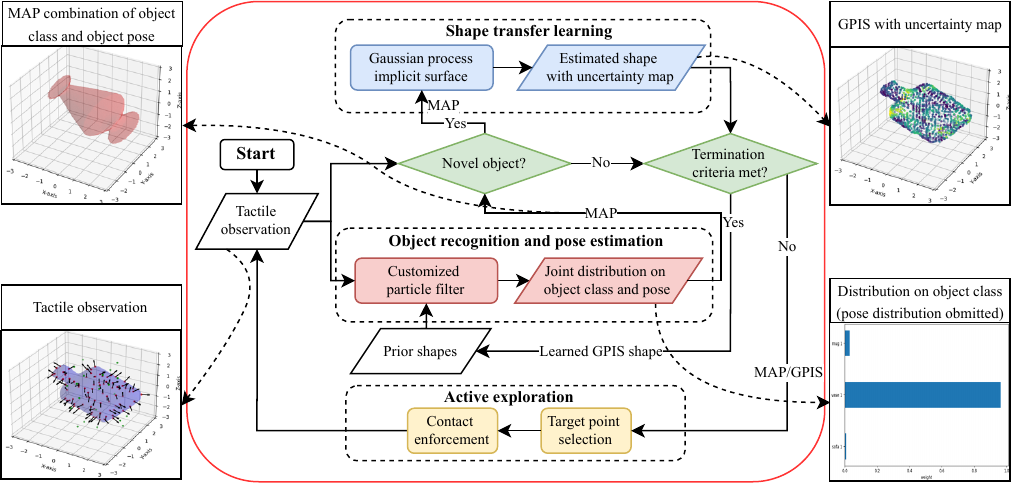}}
\caption{The proposed framework consists of a customized particle filter and a Gaussian process implicit surface (GPIS). The customized particle filter estimate the joint distribution of object pose and object class upon newly obtained tactile observation, from which the maximum a posteriori (MAP) combination of object class and pose can be extracted. The MAP is used as a prior for GPIS reconstruction when a novel object is identified. An exploration procedure based on the global shape estimation (MAP/GPIS), including target point selection and contact enforcement is proposed to perform active data acquisition. Tactile exploration continues until the termination criterion is met. The learned GPIS from a novel object can be added as a new prior, which enables the framework to recognize it in future exploration.}
\label{fig:high_level_component}
\end{figure*}
\section{Methods}
The goal of this paper is to develop a unified Bayesian framework that through active tactile exploration, differentiates between known and novel objects, estimates object class and pose for known objects, and reconstructs shapes for novel objects using the best-matching prior from known objects.
Fig.~\ref{fig:high_level_component} shows the outline and illustrates the core idea of the proposed framework. 
In this section, we describe and derive each component of the framework in detail. We start by deriving the Bayesian formulation of the object recognition and pose estimation problem for known objects with our customized particle filters, then expand the formulation to detect novel objects and incorporate GPIS to enable shape transfer learning. Lastly, explanations of the active exploration procedure based on global shape estimates and termination criterion are given.
\subsection{Bayesian Formulation of Object Recognition and Pose Estimation through Touch}

\subsubsection{Latent Variables} We denote object class as $c$ and the 6D object pose as $\mathbf{p}$. For compactness, $c$ and $\mathbf{p}$ are concatenated into a latent variable vector $\mathbf{z}$, i.e., $ \mathbf{z} \leftarrow \left [ c, \mathbf{p} \right ]$.
\subsubsection{Object Priors} Our framework encodes the knowledge of known object shapes in the form of signed distance functions (SDFs) \cite{osherSignedDistanceFunctions2003}. Given an object class $c$ and object pose $\mathbf{p}$, the predicted signed distance $\widehat{d}_s$ from a point $\mathbf{x}$ to the object surface can be obtained through the signed distance function
\begin{equation}
    \widehat{d}_s := f(\mathbf{x}, \mathbf{z}).
\end{equation}
\begin{equation}
    \begin{cases} 
      \widehat{d}_s=0 &  \text{on surface}\\
      \widehat{d}_s>0 & \text{outside object}\\
      \widehat{d}_s<0 & \text{inside object }
   \end{cases}
\end{equation}
The predicted surface normal vector $\mathbf{\widehat{n}}$ at a contact point $\mathbf{x}$ is calculated by differentiation, i.e.,
\begin{equation}
    \mathbf{\widehat{n}} = \nabla_{\mathbf{x}}{f}(\mathbf{x}, \mathbf{z}).
\end{equation}
\subsubsection{Observations} In this study, each tactile sensor is abstracted as a single point. Each tactile data point consists of the sensor location $\mathbf{x}$ in the world frame, the signed distance value $d_s$ and surface normal vector $\mathbf{n}$ observed at location $\mathbf{x}$. The observed signed distance value $d_s$ can be inferred from the contact mask, i.e.,
\begin{equation}
    \begin{cases} 
      d_s=0 &  \text{contact}\\
      d_s>0 & \text{no contact}\\
   \end{cases}.
\end{equation}
For compactness, $d_s$ and $\mathbf{n}$ is concatenated into a observation vector $\mathbf{d}$, i.e., $\mathbf{d} \leftarrow \left [d_s, \mathbf{n} \right ]$.
Similarly, the predicted value of $\mathbf{d}$ is denoted as $\mathbf{\widehat{d}}$, where $\mathbf{\widehat{d}} \leftarrow [\widehat{d}_s, \mathbf{\widehat{n}} ]$.

\subsubsection{Bayesian Inference of Object Class and Object Pose}
Let $\mathbf{D}$ denote the set of all observed tactile signals $\mathbf{d}$ and $\mathbf{X}$ be the set of all locations $\mathbf{x}$ where $\mathbf{D}$ are observed.
Given tactile sensory data $\mathbf{D}$ at locations $\mathbf{X}$, the goal is to estimate the latent variables $\mathbf{z}$. Based on Bayes' theorem, the posterior distribution on $\mathbf{z}$ can be calculated as follows,
\begin{equation}
   p(\mathbf{z}|\mathbf{D},\mathbf{X}) =  \frac{p(\mathbf{D}|\mathbf{z},\mathbf{X})p(\mathbf{z})}{\sum_{\mathbf{z}} p(\mathbf{D}, \mathbf{z}|\mathbf{X})} \propto p(\mathbf{D}|\mathbf{z},\mathbf{X})p(\mathbf{z})
   \label{eqn:posterior}
\end{equation}
where $p(\mathbf{z})$ is the prior joint distribution of object class and object pose when no data is present, which is chosen to be a uninformative uniform distribution $U$ in this study;  $p(\mathbf{D}|\mathbf{z},\mathbf{X})$ denotes the likelihood of observing tactile sensory data $\mathbf{D}$ at locations $\mathbf{X}$ given $\mathbf{z}$. The summation operation $\sum_{\mathbf{z}}$ is overloaded to represent the summation over $c$ and integrals over $\mathbf{p}$. 



As the proposed framework actively acquires data, at each time step $t$, new tactile data $\mathbf{D_t}$ are observed at locations $\mathbf{X}_t$. 
Let $T$ denote the total time steps passed, $\mathbf{D}_{1:T}$ represent all observed tactile data from time step $1$ to $T$, and $\mathbf{X}_{1:T}$ denote the locations where $\mathbf{D}_{1:T}$ are observed; then \eqref{eqn:posterior} can be rewritten as
\begin{equation}
    p(\mathbf{z} | \mathbf{D}_{1:T},\mathbf{X}_{1:T}) = \frac{p(\mathbf{z}, \mathbf{D}_{1:T}| \mathbf{X}_{1:T})}{\sum_{\mathbf{z}} p(\mathbf{z},\mathbf{D}_{1:T} | \mathbf{X}_{1:T})}.
\label{eqn:posterior_all_time}
\end{equation}
By assuming independence between different observations given $\mathbf{z}$, the joint distribution can be calculated as
\begin{equation}
   \begin{aligned}
    p(\mathbf{z}, \mathbf{D}_{1:T} | \mathbf{X}_{1:T}) &=  p(\mathbf{z})p(\mathbf{D}_{1:T-1} | \mathbf{z}, \mathbf{X}_{1:T-1})p(\mathbf{D}_{T} |\mathbf{z}, \mathbf{X}_{T})\\
    &=  p(\mathbf{z}, \mathbf{D}_{1:T-1} | \mathbf{X}_{1:T-1})  p(\mathbf{D}_{T} |\mathbf{z}, \mathbf{X}_{T}),
\label{eqn:joint_distribution}
\end{aligned} 
\end{equation}
where the recursive update rule is obtained.
Following the aforementioned independence assumption,  $p(\mathbf{D}_{t}|\mathbf{z},\mathbf{X}_{t})$ can be further decomposed into
\begin{equation}
   p(\mathbf{D}_{t}|\mathbf{z},\mathbf{X}_{t}) =\underset{i=1}{\prod^{O}} p(\mathbf{d}_{i,t}|\mathbf{z},\mathbf{x}_{i,t}).
   \label{eqn:decomposition}
\end{equation}
where the subscript $i$ denotes the index of each observed data point, and $O$ denotes the total number of observed data points. 

\subsubsection{Measurement Likelihood Function}
Assuming each tactile observation $\mathbf{d}$ at a location $\mathbf{x}$ follows a Gaussian distribution $\mathbf{d} \sim \mathcal{N}(\mathbf{\widehat{d}},\boldsymbol{\Sigma})$, then the measurement likelihood function can be written as 
\begin{equation}
   p(\mathbf{d}|\mathbf{z},\mathbf{x})  
   =\frac{1}{Z} \exp\left [-\frac{1}{2} \left (\mathbf{\widehat{d}}-\mathbf{d}\right )^T\boldsymbol{\Sigma}^{-1}\left (\mathbf{\widehat{d}}-\mathbf{d}\right )\right ],
\label{eqn:pf_likelihood}
\end{equation}
\begin{equation}
    Z = (2\pi)^{2}\sqrt{\det(\boldsymbol{\Sigma})},
\end{equation}
\begin{equation}
    \boldsymbol{\Sigma} = \text{diag}\left(\left[\sigma_d^2, \sigma_n^2,\sigma_n^2,\sigma_n^2 \right ]\right),
\end{equation}
where $Z$ is the normalization constant, diag is an operator to form a diagonal matrix from a vector, $\sigma_d$ and $\sigma_n$ denote the standard deviations for the signed distance observation and surface normal vector observation, respectively.

 \subsubsection{Negative Information Update}
 For non-contact observations, the likelihood integrates over $ds > 0$, it follows,
 \begin{equation}
 p(\mathbf{d}|\mathbf{z},\mathbf{x}) = p(d_{s} > 0|\mathbf{z},\mathbf{x}) = \int_{0}^{\infty} p(d_{s} |\mathbf{z},\mathbf{x}) d(d_{s}).  
 \label{eqn:int}
 \end{equation}
 For the Gaussian likelihood function, the likelihood reads  
 \begin{equation}
     p(d_{s} > 0|\mathbf{z},\mathbf{x}) = \frac{1}{2} \left(1 - \mathbf{erf}\left( \frac{ - f(\mathbf{x},\mathbf{z})}{\sqrt{2}\sigma_d}\right)\right).
     \label{eqn:negative}
 \end{equation}
where $\mathbf{erf}$ denotes the Gauss error function.
By substituting \eqref{eqn:int}\eqref{eqn:negative} into \eqref{eqn:decomposition}, the framework can update its belief using non-contact points.

\subsubsection{MAP Model Evidence}
The MAP combination of the object pose and object class, denoted by $\mathbf{z^*}$, is obtained by,
\begin{equation}
    \mathbf{z^*} := \argmax_{\mathbf{z}}( p(\mathbf{z} | \mathbf{D}_{1:T}, \mathbf{X}_{1:T})).\\
\label{eqn:mle_obj}
\end{equation}

The framework tracks the fitness between the $\mathbf{z}^*$ and the observed data $\mathbf{D}_{1:T}$ by calculating $ p(\mathbf{D}_{1:T} | \mathbf{z}^*, \mathbf{X}_{1:T})$ using  \eqref{eqn:decomposition}. 
In the following sections, $p(\mathbf{D}_{1:T} | \mathbf{z}^*, \mathbf{X}_{1:T})$ is referred to as MAP model evidence. 

\subsection{Customized Particle Filter for Object Recognition and Pose Estimation}
Combining \eqref{eqn:posterior_all_time}-\eqref{eqn:negative}, one can update the posterior distribution $p(\mathbf{z}|\mathbf{D}_{1:t},\mathbf{X}_{1:t})$ recursively for each time step $t$ in theory. However, in practice the term $\sum_{\mathbf{z}} p(\mathbf{z},\mathbf{D}_{1:T} | \mathbf{X}_{1:T})$ in \eqref{eqn:posterior_all_time} is intractable. To address this issue, 
a particle filter (PF) is used to approximate the posterior distribution of interest with weighted discrete samples (particles) based on importance sampling, i.e.,
\begin{equation}
    p(\mathbf{z} | \mathbf{D}_{1:t},\mathbf{X}_{1:t}) = \sum_{j=1}^{N} \overline{w}_{j,t} \; \delta(\mathbf{z}-\mathbf{z}_{j,t}),
    \label{eqn:importance_sampling}
\end{equation}
where the subscript $j,t$ denotes the index of a particle in the PF at time step $t$, $\overline{w}_{j,t}$ denotes the normalized weight of the particle $\mathbf{z}_{j,t}$, $\delta$ denotes the Dirac delta function, and $N$ denotes the total number of particles.
With the observational independence assumption, the unnormalized weight $w_{j,t}$ of the particle $\mathbf{z}_{j,t}$ can be calculated as
\begin{equation}
\begin{aligned}
        w_{j,t} &\propto p(\mathbf{z}_{j,t} | \mathbf{D}_{1:t},\mathbf{X}_{1:t}) \\
        &\propto p(\mathbf{z}_{j,t}, \mathbf{D}_t | \mathbf{D}_{1:t-1},\mathbf{X}_{1:t})\\
        &= p(\mathbf{z}_{j,t} | \mathbf{D}_{1:t-1},\mathbf{X}_{1:t})\; p( \mathbf{D}_{t} | \mathbf{z}_{j,t}, \mathbf{D}_{1:t-1},\mathbf{X}_{1:t})\\
        & = p(\mathbf{z}_{j,t} | \mathbf{D}_{1:t-1},\mathbf{X}_{1:t-1})\; p( \mathbf{D}_{t} | \mathbf{z}_{j,t}, \mathbf{X}_{t})\\
        &= \sum_{\mathbf{z}_{j,t-1}} (\overline{w}_{j,t-1}\; p(\mathbf{z}_{j,t} | \mathbf{z}_{j,t-1},\mathbf{D}_{1:t-1},\mathbf{X}_{1:t-1})) \times \\ 
        &\; \; \; \; \; \; \;  p( \mathbf{D}_{t} | \mathbf{z}_{j,t}, \mathbf{X}_{t}).
\end{aligned}
\label{eqn:importance_weight}
\end{equation}

Since we assume the object stays static during the exploration, thus
\begin{equation}
    \begin{aligned}
           p(\mathbf{z}_{j,t} | \mathbf{z}_{j,t-1},\mathbf{D}_{1:t-1},\mathbf{X}_{1:t-1})
           &= \delta(\mathbf{z}_{j,t}-\mathbf{z}_{j,t-1}),
    \end{aligned}
    \label{eqn:static_assumption_1}
\end{equation}
\begin{equation}
    \mathbf{z}_{j,t} = \mathbf{z}_{j,t-1}.
    \label{eqn:static_assumption_2}
\end{equation}

Noteworthy, $p(\mathbf{z}_{j,t} | \mathbf{z}_{j,t-1},\mathbf{D}_{1:t-1},\mathbf{X}_{1:t-1})$ can be replaced by a motion estimator if a motion model is available. With \eqref{eqn:static_assumption_1} and \eqref{eqn:static_assumption_2},
it follows
\begin{equation}
    \begin{aligned}
        w_{j,t} &= \sum_{\mathbf{z}_{j,t-1}} (\overline{w}_{j,t-1}\; \delta(\mathbf{z}_{j,t}-\mathbf{z}_{j,t-1})) \;  p( \mathbf{D}_{t} | \mathbf{z}_{j,t}, \mathbf{X}_{t})\\
        &= \overline{w}_{j,t-1}\; p( \mathbf{D}_{t} | \mathbf{z}_{j,t}, \mathbf{X}_{t}),
    \end{aligned}
    \label{eqn:importance_weight_unnormalized}
\end{equation}
\begin{equation}
    \overline{w}_{j,t} = \frac{w_{j,t}}{\sum_{j=1}^{N} w_{j,t}}.
    \label{eqn:importance_weight_normalized}
\end{equation}
where $\overline{w}_{j,t}$ is the normalized weight of particle $\mathbf{z}_{j,t}$. 
Using \eqref{eqn:importance_weight_normalized} and \eqref{eqn:importance_weight_normalized}, one can update each particle's weight recursively at each time step.

As the particle filter needs to estimate the joint distribution of the object's class and its 6-DOF pose, the number of particles required to cover the space sufficiently can be calculated as $n\times (m^6)$, where $n$ denotes the number of known object classes, and $m$ denotes the resolution of the discretization at each continuous dimension.

To enhance the sample efficiency of the particle filter and have better coverage on the high-density region of the posterior distribution, we propose to sample new particles based on newly observed data at each time step and mix them with existing particles.
The rotation and translation invariant point-pair feature proposed in  \cite{drostModelGloballyMatch2010} is adopted for this purpose. For every two data points, the point-pair feature can be calculated.
\begin{figure}[!t]
\centerline{\includegraphics[width = 1.0\linewidth]{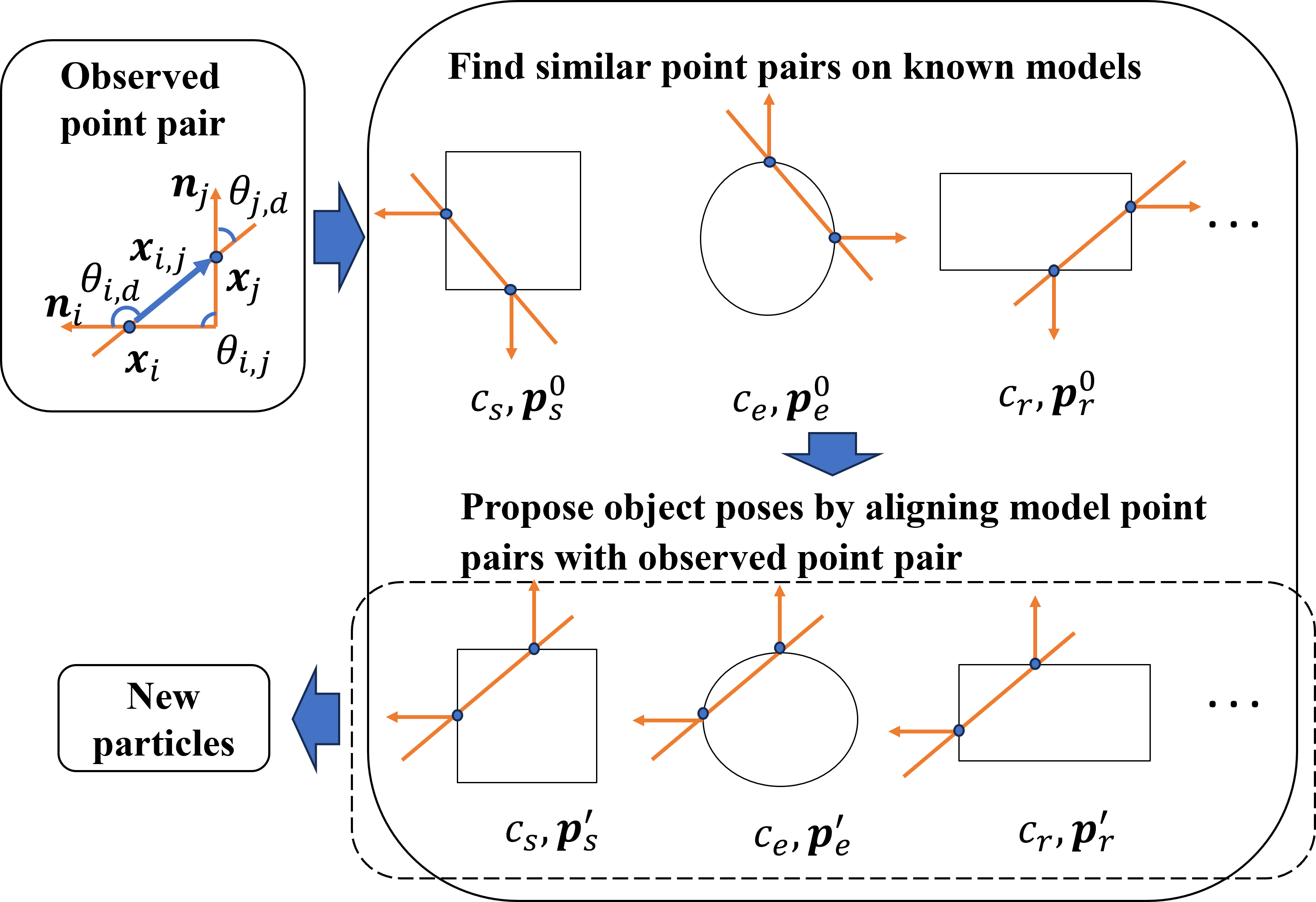}}
\caption{An example of sampling particles with point-pair features. For each observed data point pair, $\theta_{i,j},\theta_{i,d},\theta_{j,d}$ and $||\mathbf{x}_{i,j}||_2$ are calculated as the point pair features. $\mathbf{x}_{i,j}$ denotes the vector $\mathbf{x}_j-\mathbf{x}_i$. 
$\theta_{i,j}$ ,$\theta_{i,d}$, $\theta_{j,d}$ are the angle between normal vectors of $\mathbf{n}_i$ and $\mathbf{n}_j$, the angle between $\mathbf{n}_i$, and $\mathbf{x}_{i,j}$, the angles between $\mathbf{n}_j$ and $\mathbf{x}_{i,j}$ respectively. Point pairs on all known models with similar features are extracted. Finally, by aligning known model point pairs with the observed point pair, possible combinations of the object class and object pose that match the observed point pair are found. $c_s$, $c_e$, $c_r$ represent three object classes, namely square, ellipse, and rectangular respectively.  $\mathbf{p_{s}^{0}}$, $\mathbf{p_{e}^{0}}$, $\mathbf{p_{r}^{0}}$ are the original object pose, by convention set to $[0,0,0,0,0,0]$, whereas $\mathbf{p_{s}^{'}}$, $\mathbf{p_{e}^{'}}$, $\mathbf{p_{r}^{'}}$ are the pose after the alignment.}
\label{fig:pair_wise_proposal}
\end{figure}
Fig.~\ref{fig:pair_wise_proposal} illustrates the sampling procedure in a simplified 2D case. The same procedure holds for the 3D case, with more sophisticated objects.

For each known object, 200 feature points are sampled using Poisson Disk Sampling, resulting in $200*199=39800$ point pairs. The point-pair features for point pairs on known models are pre-computed, discretized and stored in a hash table for quick look-up at runtime. 
Upon a newly obtained contact point, point-pair features between the new contact point and previous contact points are calculated, discretized, and the hash table is used to retrieve corresponding point pairs on known object models with the same point-pair feature values.
The discretization is done by rounding $||\mathbf{x}_{i,j}||_2$ to the closest one digit decimal and $\theta_{i,j}$ ,$\theta_{i,d}$, $\theta_{j,d}$ to the closest multiple of $12$ degrees. 
The more coarse the discretization is, the more forgiving the sampling scheme is, this would propose particles even when larger noise is present, but undermining sampling efficiency. 
The reported discretization resolution is chosen empirically for the (low-noise) simulated experiments. 

Once the correspondences between observed point pairs and model point pairs are found, aligning the model point pairs with newly observed point pairs yields hypotheses of object class and object pose that at least match one observed contact point pair. 
A unique transformation can be derived to align two point pairs except for degenerate bilinear cases.
For the alignment process, see \cite{drostModelGloballyMatch2010}.

The look-up table and alignment processes together can be viewed as sampling particles from the mixture of posterior distributions $p(\mathbf{z}|\mathbf{x}_i,\mathbf{n}_i,\mathbf{x}_j,\mathbf{n}_j)$ conditioned on each observed point pair. This sampling procedure can concentrate new particles in high probability density regions of the true posterior distribution which the PF aims to approximate. Note that early ruled out states can be sampled again if they match the new point pairs, which is valuable when trying to find the best prior for an unknown object, since an overall good prior can be pruned out earlier due to mismatch with partial observations.

As tactile data points are obtained sequentially in this study, at time step $t$, pairing the new contact with all $t-1$ previous contacts leads to $2(t-1)$ ordered pairs, which means that sampling new particles becomes computationally expensive as $t$ increases. Additionally, to assign proper weights to new particles, it is costly to evaluate all new particles on all previous observations with large $t$. 

To keep the computation tractable, a subset $\mathbf{X}_s \subset \mathbf{X}_{1:t-1}$ of  a fixed number $n_s$ of previous contact locations is selected at each time step. Previous contacts are sorted by distance to the new contact and divided into $n_s$ equally sized segments and the first point within each segment constitutes $\mathbf{X}_s$. The tactile data observed at $\mathbf{X}_s$ is denoted as $\mathbf{D}_s$. Only $\mathbf{X}_s$ and $\mathbf{D}_s$ are used to construct $2n_s$ ordered pairs of points for sampling and to compute relative particle weights among the new particles. This strategy maintains a near fixed computational cost per step while capturing better global spatial context than a temporal sliding-window-based approach \cite{vezzaniMemoryUnscentedParticle2017}.


The derivation of the weight approximation scheme goes as follows. Let us denote the weight of new particles $\mathbf{z}^{'}_{c,k,t}$ with class $c$ by $w^{'}_{c,k,t}$ and the highest weight among them by $w^{'}_{c,*,t}$. The ratio between $w^{'}_{c,k,t}$ and $w^{'}_{c,*,t}$ is determined by:
\begin{equation}
    \frac{w^{'}_{c,k,t}}{w^{'}_{c,*,t}} = \frac{p(\mathbf{D}_{s}| \mathbf{z}^{'}_{c,k,t}, \mathbf{X}_{s})}{p(\mathbf{D}_{s}| \mathbf{z}^{'}_{c,*,t}, \mathbf{X}_{s})}.
    \label{eqn:relative_weight_1}
\end{equation}
For each object, class $c$, $\mathbf{z}^{'}_{c,k,t}$ will be evaluated based on all previous observations $\mathbf{D}_{1:t}$ at locations $\mathbf{X}_{1:t}$ to determine its weight $w^{'}_{c,*,t}$ with respect to the weight $\overline{w}^{*}_{t}$ of the MAP particle $\mathbf{z}^{*}_{t}$ at time step $t$. In other words,
\begin{equation}
    \frac{w^{'}_{c,*,t}}{\overline{w}^{*}_{t}} = \frac{p(\mathbf{D}_{1:t}| \mathbf{z}^{'}_{c,*,t}, \mathbf{X}_{1:t})}{p(\mathbf{D}_{1:t}| \mathbf{z}^{*}_{t}, \mathbf{X}_{1:t})}
    \label{eqn:relative_weight_2}
\end{equation}
Combining \eqref{eqn:relative_weight_1} and \eqref{eqn:relative_weight_2}, 
\begin{equation}
    w^{'}_{c,k,t} = \frac{p(\mathbf{D}_{s}| \mathbf{z}^{'}_{c,k,t}, \mathbf{X}_{s})}{p(\mathbf{D}_{s}| \mathbf{z}^{'}_{c,*,t}, \mathbf{X}_{s})}\frac{p(\mathbf{D}_{1:t}| \mathbf{z}^{'}_{c,*,t}, \mathbf{X}_{1:t})}{p(\mathbf{D}_{1:t}| \mathbf{z}^{*}_{t}, \mathbf{X}_{1:t})}\overline{w}^{*}_{t}
    \label{eqn:new_weights}
\end{equation}

One key characteristic of this approximation is that $w^{'}_{c,k,t}$ can be higher than $\overline{w}^{*}_{t}$ if and only if $\mathbf{z}^{'}_{c,*,t}$ is a better fit compared to $\mathbf{z}^{*}_{t}$ over all observations $\mathbf{D}_{1:t}$. 

To further reduce the computational cost, the proposed sampling scheme is only performed when the MAP model evidence $p(\mathbf{D}_{1:t} | \mathbf{z}^*, \mathbf{X}_{1:t})$ is smaller than a certain threshold $\lambda$. In other words, if the MAP is a good fit for the data, new particles will not be proposed.

The pseudo-code for the PF is shown in Algorithm \ref{alg:obj}. At each time step, a resampling step is carried out before sampling new particles from point-pair features to keep the total number of particles small. In this study, the stochastic universal sampling (SUS) is used. The SUS guarantees the survival of particles with normalized weights larger than $\frac{1}{N}$, when $N$ is the number of samples to be drawn.\\

\begin{algorithm}[ht]
\label{alg:obj}
\caption{Particle Filter (PF) for Object Recognition and Pose Estimation }
\begin{algorithmic}
\STATE $\mathbf{z}_{j,t}$ := class and the pose of particle $j$ at time step $t$
\STATE $\overline{w}_{j,t}$ :=  normalized weight of particle $j$ at time step $t$
\STATE $D_{t}$ := the tactile data observed at time step $t$
\STATE $X_{t}$ := the data point locations at time step $t$
\STATE $T$ := total number of time steps
\STATE $S$ := the set of all particles in the PF
\STATE \textbf{Initialization, when $t=1$:}
\STATE Sample particles by aligning each feature point and its normal vector on known models with the first oriented contact point.
\STATE $ \mathbf{S} \leftarrow {((\mathbf{z}_{1,1}, \overline{w}_{1,1}), \cdots, (\mathbf{z}_{N,1}, \overline{w}_{N,1}))}$


\FOR{$t=2,\cdots,T$}
    \FOR{$(\mathbf{z}_{j,t},\overline{w}_{j,t-1}) \in \mathbf{S}$}
        \STATE \textbf{Weight update:} Update the normalized weight $\overline{w}_{j,t-1}$ to $\overline{w}_{j,t}$ for particle $j$ based on \eqref{eqn:importance_weight_unnormalized} and \eqref{eqn:importance_weight_normalized}
    \ENDFOR
    \STATE \textbf{Resampling:} 
    $S \leftarrow$ \text{Stochastic Universal Sampling}(S)
    \STATE \textbf{Tracking model evidence:} Find the MAP particle $(\mathbf{z}^{*}_{t},\overline{w}^{*}_{t})$ based on \eqref{eqn:mle_obj} and calculate the model evidence $p(\mathbf{D}_{1:t} | \mathbf{z}^*_{t}, \mathbf{X}_{1:t})$
    \IF {$p(\mathbf{D}_{1:t} | \mathbf{z}^*_{t}, \mathbf{X}_{1:t}) \leq \lambda$}
        \STATE \textbf{Sampling particles:} Sample new particles $(\mathbf{z}^{'}_{k,t},w^{'}_{k,t})$ based on point-pair features of observed point pairs and  \eqref{eqn:relative_weight_1}-\eqref{eqn:new_weights}, in total $K$ particles are sampled.
    \FOR {$k=1,\cdots,K$}
        \STATE$ \mathbf{S} \leftarrow \mathbf{S} \cup (\mathbf{z}^{'}_{k,t},w^{'}_{k,t})$
    \ENDFOR
    \ENDIF
\ENDFOR
\end{algorithmic}

\end{algorithm}


\subsection{Discriminating between Known and Novel Objects}
To discriminate between known and novel objects, a threshold needs to be defined for the MAP model evidence $ p(\mathbf{D}| \mathbf{z}^*, \mathbf{X})$.
Since both contact points and non-contact points can be present, the criterion for an object to be classified as a known object is defined as follows
\begin{equation}
   p(\mathbf{D} | \mathbf{z}^*, \mathbf{X}) \geq \frac{1}{Z}(0.9)^{n_{pos}} * (0.5)^{n_{neg}},
    \label{eqn:mle_thershold}
\end{equation}
where $n_{pos}$ and $n_{neg}$ denote the numbers of observed contact and non-contact points, respectively. From \eqref{eqn:pf_likelihood}, the maximal likelihood of an exact contact is equal to $\frac{1}{Z}$. The threshold therefore encourages the average likelihood of contact observations to exceed $90\%$ of this maximum (close to the MAP surface), while non-contact observations should have likelihood at least $0.5$ (no penetration with the MAP surface). The choice of threshold and its impact are discussed further in the discussion section.

\subsection{Gaussian Process Implicit Surface for Shape Reconstruction}
Once a novel object is detected, the Gaussian Process Implicit Surface (GPIS) \cite{williamsGaussianProcessImplicit2006} is used to reconstruct the shape given a prior function $\boldsymbol{\mu}$ and observed data $\mathbf{D}$. The aim is to learn a signed distance function that fits the data $\mathbf{D}$ while taking into account $\boldsymbol{\mu}$, where $\boldsymbol{\mu}: \mathbb{R}^3 \rightarrow \mathbb{R}^4$ maps a point $\mathbf{x}$ to a predicted signed distance value and a gradient vector.

Under the GP assumption, given an unexplored location $\mathbf{x}^*$, observed contact points $\mathbf{X}$ and observed tactile data $\mathbf{D}$, the predictive observation $\mathbf{d}^*$ at $\mathbf{x}^*$ satisfies,
\begin{equation}
    \mathbf{d}^{*} \sim N(\boldsymbol{\mu}_{p}(\mathbf{x}^*),\boldsymbol{\Sigma}_{p}(\mathbf{x}^*)).
\end{equation}
The mean $\boldsymbol{\mu}_{p}(\mathbf{x}^*)$ and the covariance matrix $\boldsymbol{\Sigma}_{p}(\mathbf{x}^*)$ are given by

\begin{align}
    \boldsymbol{\mu}_{p}(\mathbf{x}^*) &= \boldsymbol{\mu}(\mathbf{x}^*)+\mathbf{k}_{*}(\mathbf{K}+\sigma^2 \mathbf{I})^{-1}(\mathbf{\overline{D}}-\overline{\boldsymbol{\mu}}(\mathbf{X})), \\
    \boldsymbol{\Sigma}_{p}(\mathbf{x}^*) &= \mathbf{k}_{**}-\mathbf{k}_{*}(\mathbf{K}+\sigma^2 \mathbf{I})^{-1}\mathbf{k}_{*}^T
    \label{posterior_variance}
\end{align}
where $\sigma$ denotes the sensory noise level; $\mathbf{\overline{D}}$ and $\overline{\boldsymbol{\mu}}(\mathbf{X})$ represents the flattened vectors of $\mathbf{D}$ and $\boldsymbol{\mu}(\mathbf{X})$ , respectively; $\mathbf{k}_{**}$, $\mathbf{k}_{*}$ and $\mathbf{K}$ is the covariance matrix between $(\mathbf{d}^*, \mathbf{d}^*)$, $(\mathbf{d}^*, \overline{\mathbf{D}})$ and $(\overline{\mathbf{D}},\overline{\mathbf{D}})$ respectively. A covariance matrix between observation $\mathbf{d}$ at $\mathbf{x}$ and observation $\mathbf{d}^{'}$ at $\mathbf{x}^{'}$ are then given by
\begin{equation}
   cov(\mathbf{d},\mathbf{d^{'}}) = 
   \begin{bmatrix}   k_f(\mathbf{x},\mathbf{x^{'}}) & \frac{\partial}{\partial\mathbf{x}^{'}}\left (k_f(\mathbf{x},\mathbf{x}^{'}) \right ) \\
   \frac{\partial^{T}}{\partial\mathbf{x}} \left (k_f(\mathbf{x},\mathbf{x^{'}}) \right) & \frac{\partial^{T}}{\partial\mathbf{x}^{'}} \frac{\partial}{\partial\mathbf{x}}\left (k_f(\mathbf{x},\mathbf{x^{'}})\right)\\
    \end{bmatrix},
    \label{eqn:gp_cov}
\end{equation}
where $k_f$ is the kernel function of choice. Following \eqref{eqn:gp_cov}, $\mathbf{k}_{**}$, $\mathbf{k}_{*}$ and $\mathbf{K}$ can be calculated.
 
 In this study, the thin-plate kernel \cite{williamsGaussianProcessImplicit2006} \cite{martensGeometricPriorsGaussian2017} is used. 
The thin-plate kernel function is defined as follows,
\begin{equation}
    k_{f}(\mathbf{x},\mathbf{x}^{'}) = k_{TP}(\mathbf{x},\mathbf{x}^{'}) = a(2d^3 - 3Rd^2 + R^3) 
\end{equation}
\begin{equation}
    d =||\mathbf{x}-\mathbf{x}^{'}||_2
\end{equation}
The first and second derivatives of $k_{TP}(\mathbf{x},\mathbf{x}^{'})$ can be calculated as
\begin{align}
    \frac{\partial{k_{TP}}(\mathbf{x},\mathbf{x^{'}})}{\partial{x_i}} &= 6a(x_i-x^{'}_i)(d-R), \\ 
    \frac{\partial^2{k_{TP}}(\mathbf{x},\mathbf{x^{'}})}{\partial{x_i}\partial{x^{'}_j}} &= -6a\left(\frac{(x_i-x^{'}_i)(x_j-x^{'}_j)}{d} + \delta_{ij}(d-R)\right),
\end{align}
where $\delta_{ij}$ denotes the Dirac delta function.
 Differing from the definition in \cite{williamsGaussianProcessImplicit2006}\cite{martensGeometricPriorsGaussian2017}, a scaling coefficient $a$ is introduced to allow the GP to adapt to various novel objects. Larger $a$ corresponds to assuming larger differences between the prior and the novel object. 
 The kernel parameter $a$ is updated online using one-step Newton's method to maximize the data likelihood of the GP \cite{murphyProbabilisticMachineLearning2022}. Last but not least, only contact points are used to update the GP due to inexact signed distance value observation at non-contact points.

\subsection{Combining PF and GPIS}
The GPIS provides the ability of adapting to, and learning, novel shapes, whereas the PF can find the maximum likelihood combination of known object class and pose based on the observed data. Therefore, we propose to use the MAP particles  of the PF as a prior for GPIS shape reconstruction.
In this manner, the GPIS has a more flexible, grounded prior selection process and the knowledge from known shapes can be transferred to learn new shapes.

For GPIS reconstruction, given a fixed prior function $\boldsymbol{\mu}$ based on the MAP estimation $\mathbf{z}^*$ from the PF, the likelihood of observing data $\mathbf{D}$ at a set of points $\mathbf{X}$ satisfies the following relation:
\begin{equation}
\begin{aligned}
    p(\mathbf{D} | \mathbf{z}^*, \mathbf{X}) \propto \exp \left [\left (\overline{\boldsymbol{\mu}}(\mathbf{z}^*,\mathbf{X})- \overline{\mathbf{D}}\right )^T \mathbf{K}^{-1} \left (\overline{\boldsymbol{\mu}}(\mathbf{z}^*,\mathbf{X})-\overline{\mathbf{D}}\right ) \right ].
\end{aligned}
\label{eqn:log_gp_1}
\end{equation}
Notice the likelihood function from \eqref{eqn:log_gp_1} takes a similar form to \eqref{eqn:pf_likelihood}. The main difference is that the diagonal matrix $-\frac{1}{2}\boldsymbol{\Sigma}^{-1}$ is replaced with the inverse of the kernel matrix $\mathbf{K}^{-1}$, which hints the relation between the two formulations.
In the PF, there is no spatial correlation among data points with the known object class assumption, hence the diagonal matrix. While for novel object, spatial correlation from the GPIS's smoothness assumption is necessary for inferring the signed distance values and the surface normal vectors at unexplored locations. Our formulation utilizes both prior knowledge on known objects and the observed data for prediction.

\subsection{Active Data Acquisition}
 Due to the locality of tactile observations, active data acquisition is required to address the uncertainty and ambiguity in object identity, pose and shape. The active data acquisition procedure is divided into two steps: target point selection and contact enforcement. 

 \subsubsection{Target Point Selection}
When the object is classified as novel, the point with the maximal posterior variance on the GPIS is selected as the next target point, i.e.,\\
\begin{equation}
    \overline{\mathbf{x}}_{t+1} = \argmax_{\mathbf{x}^* \in \mathbf{S}^*}  \left \{\text{var}(\mathbf{x}^*) \right \},
\label{eqn:active_gp}
\end{equation}
\begin{equation}
    \text{var}(\mathbf{x}^*) = \boldsymbol{\Sigma}_{p}(\mathbf{x}^*)_{0,0},
\end{equation}
\begin{equation}
    \mathbf{S}^* \subset \left \{  \mathbf{x}^* \in R^3 \; |\; \boldsymbol{\mu}_{p}(\mathbf{x}^*) = 0   \right \}, 
\end{equation}
with $\mathbf{S}^*$ a finite subset of the zero-level set of the posterior GPIS, obtained through the marching cubes algorithm \cite{lewinerEfficientImplementationMarching2003}, and $\boldsymbol{\Sigma}_{p}(\mathbf{x}^*)_{0,0}$ denotes the first element of $\boldsymbol{\Sigma}_{p}(\mathbf{x}^*)$ in \eqref{posterior_variance}.



When the object is classified as known, GPIS updates are omitted for efficiency and the MAP surface from the PF is used instead. In this case, exploration is guided by the directed Hausdorff distance (DHD), which is defined as:
\begin{equation}
    \mathbf{d}_{H}(\mathbf{A},\mathbf{B}) = \max_{\mathbf{a} \in \mathbf{A}}  \min_{\mathbf{b} \in \mathbf{B}}\lVert \mathbf{a} - \mathbf{b} \rVert_2
\label{eqn:dhd}
\end{equation}
where $\mathbf{A}$ and $\mathbf{B}$ are two point sets.
The point on the MAP surface with the largest distance to its nearest data point is selected as the next target point:
\begin{equation}
    \overline{\mathbf{x}}_{t+1} =  \argmax_{\mathbf{x}_{m} \in \mathbf{M}}  \min_{\mathbf{x}_{c} \in \mathbf{X}_{c}}\lVert \mathbf{x}_{m} - \mathbf{x}_{c} \rVert_2,
\label{eqn:exploration_DHD}
\end{equation}
where $\mathbf{M}$ and $\mathbf{X}_{c}$ represent the sets of vertices of the MAP shape and the observed contact points, respectively. 

Both GPIS-based and DHD-based target point selection schemes prioritize surface regions far from existing data, encouraging uniform coverage of the estimated surface. Together, they are referred to as the GPIS-DHD exploration procedure in the following sections.

\subsubsection{Contact Enforcement}

\begin{figure}[!t]
\centerline{\includegraphics[width = \linewidth]{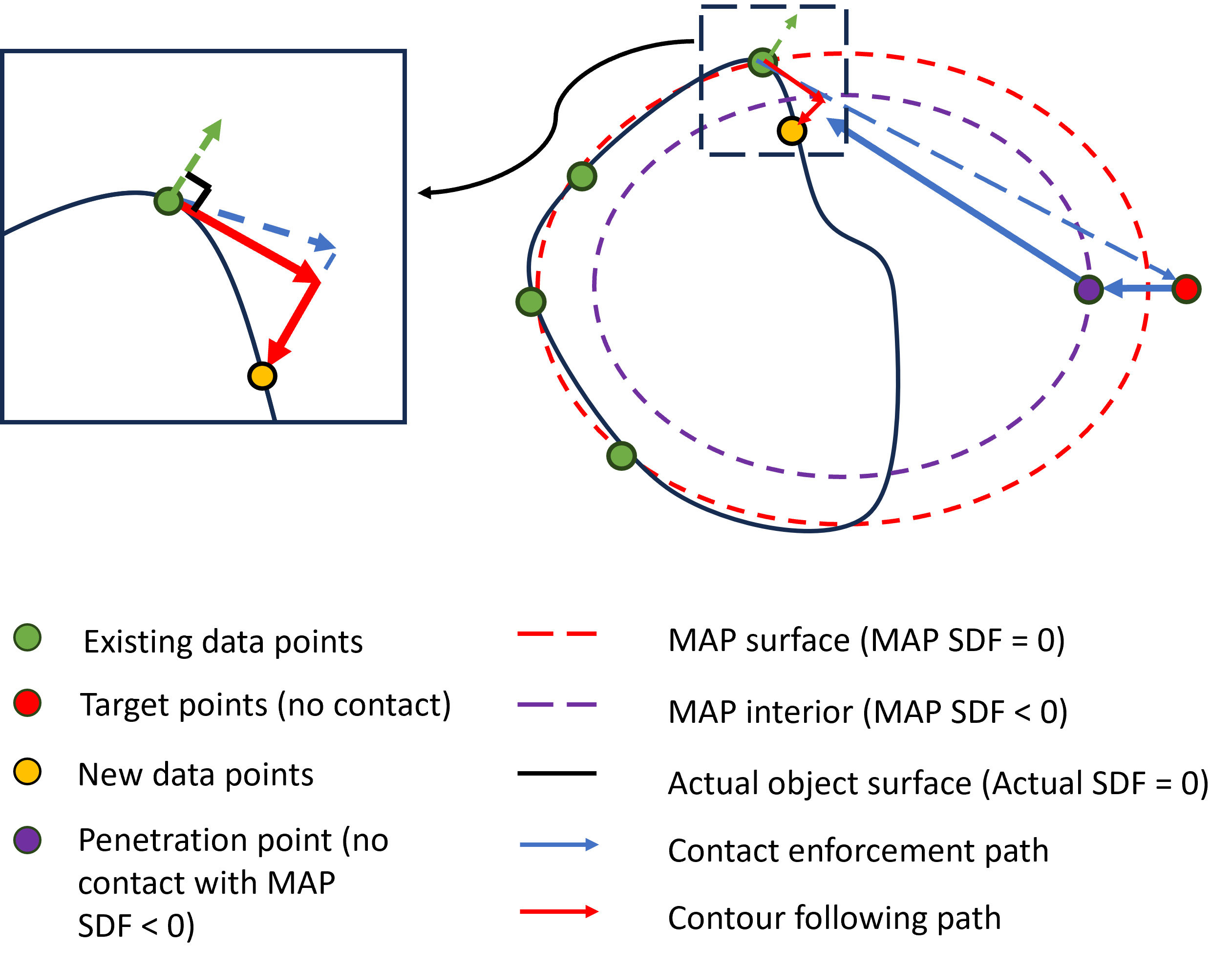}}
\caption{Example of the contact enforcement procedure. Starting from the target point (red dot), the sensor first moves towards the interior of the MAP shape (purple dot). If no contact is found, a non contact point is recorded and the sensor moves towards the closest existing contact point (green dot). If no contact is established either, the closest known contact point is contacted again. From there, the algorithm takes small steps on the surface along the local tangent plane, towards the target point while remaining in contact with the surface. After a short distance, a new contact point will ultimately be recorded (yellow dot).}
\label{fig:contact_enforcement}
\end{figure}

A selected target point does not guarantee contact. The sensor therefore needs to establish contact through an enforcement procedure as shown in Fig.~\ref{fig:contact_enforcement}: it first moves along the negative signed-distance gradient of the corresponding SDF of the estimated surface (MAP/GPIS) up to a certain depth into the object surface; if no contact is detected, a non-contact point is recorded and it moves toward the nearest observed contact point and performs local surface following towards the target region. The non-contact point plays a pivotal role in rejecting false hypotheses.

Alternatively, when no prior surface estimate is available, a fallback exploration based on a rapidly exploring random tree (RRT) is employed. A random workspace sample is projected onto the tangent plane at its nearest contact point, and a small surface-constrained step is executed towards the random sample. This local strategy does not require a global shape estimate and is used as a baseline for comparison with the GPIS-DHD procedure.

\subsection{Termination Criterion}
Exploration terminates when surface coverage reaches a pre-defined density $\epsilon$ measured by the directed Hausdorff distance.
For known objects, the termination criterion can be written as
\begin{equation}
    \mathbf{d}_{H}(\mathbf{M},\mathbf{X}_c) \leq \epsilon.
\end{equation}
Similarly, for novel objects, the termination criterion is
\begin{equation}
    \mathbf{d}_{H}(\mathbf{\mathbf{S}^*},\mathbf{X}_c) \leq \epsilon.
\end{equation}
Although the posterior variance of the GPIS can also be used for the same purpose, the proposed criterion $\mathbf{d}_{H}(\mathbf{\mathbf{S}^*},\mathbf{X}_c)$ has the advantage of being independent of the choice of kernel parameters and the type of test objects at hand.\\
Intuitively, if any point on the estimated surface is at most $\epsilon$ away from its closest existing contact point, the program terminates, i.e., the contact points have to cover the estimated object surface with a certain density related to $\epsilon$. The threshold $\epsilon$ can be interpreted as a level of detail parameter: lower $\epsilon$ results in denser data coverage on the estimated surface at the cost of longer exploration time.
With all the components defined, the pseudo code for the active tactile perception framework is given in Algorithm~\ref{alg:framework}.


\begin{algorithm}[ht]
\caption{Bayesian framework for active object recognition, pose estimation and shape reconstruction}
\label{alg:framework}
\begin{algorithmic}
\STATE \textbf{Initialization}: Initialize the particle filter with the first contact point $\mathbf{x}_{1}$ and tactile observation $\mathbf{d}_{1}$.\\
\WHILE{$\mathbf{d}_{H}(\mathbf{\mathbf{S}^*},\mathbf{X}_c) > \epsilon$}
    \STATE Update belief using the particle filter, extract the MAP particle $(\mathbf{z}^{*},\overline{w}^{*})$ and its model evidence $p(\mathbf{D}| \mathbf{z}^{*}, \mathbf{X})$
    \IF{$p(\mathbf{D}| \mathbf{z}^{*}, \mathbf{X}) < (\frac{1}{Z}0.9)^{n_{pos}} * (0.5)^{n_{neg}}$}
        \STATE Update GPIS and select the next target point $\overline{\mathbf{x}}_{t+1}$ from the zero level set of GPIS $\mathbf{S}^{*}$ using  \eqref{eqn:active_gp}
    \ELSE
        \STATE Select the next target point $\overline{\mathbf{x}}_{t+1}$ from the set of MAP surface $\mathbf{M}$ using \eqref{eqn:exploration_DHD}
    \ENDIF
    \STATE Establish contact using the exploration procedure and add observed location to $\mathbf{X}$ and tactile observation to $\mathbf{D}$.\\
    \STATE Update $\mathbf{d}_{H}(\mathbf{\mathbf{S}^*},\mathbf{X}_c)$ based on \eqref{eqn:dhd}
\ENDWHILE
\end{algorithmic}
\end{algorithm}

\subsection{Shape Similarity Metric}
In this study, the two-way Hausdorff distance (TWD) is used to measure the difference between two shapes:
\begin{equation}
    \mathbf{d}_{TH}(\mathbf{A},\mathbf{B}) = \max(\mathbf{d}_{H}(\mathbf{A},\mathbf{B}), \mathbf{d}_{H}(\mathbf{B},\mathbf{A}))
\end{equation}
The pose estimation error is measured by the TWD between the MAP shape from the PF and the actual shape of the test object. 
Similarly, the shape reconstruction error is measured by the TWD between the reconstructed shape from the GPIS and the actual shape of the test object.

We choose the TWD as it is sensitive to the miss-matching parts (the worst case) of two shapes. This is beneficial for capturing and analyzing error which can be otherwise difficult to detect with an average performance metric.

\section{Experiment Setup}

\subsection{Simulation Environment}

Experiments are conducted using a signed-distance-field-based contact simulator. The tactile sensor is modeled as a point. Given its 3D position and the ground-truth SDF, contact is detected when the signed distance value crosses zero. The sensor follows the exploration trajectory with a step size of $1/300$ of the maximal object length. Contact points are computed via bisection at zero-crossings. If the senor's initial position is inside the object, it would follow the ground truth SDF gradient to the surface and record a contact point.

\subsection{Dataset}
\begin{figure}[!ht]
\centerline{\includegraphics[width = \linewidth]{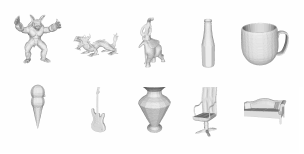}}
\caption{The ten known objects used in the experiments. From left to right, top to bottom, the objects are named as armadillo, asian dragon, elephant, bottle 1, mug 1, ice cream 1, guitar 1, vase 1, office chair and sofa 1 respectively.}
\label{fig:known}
\end{figure}
\begin{figure}[!ht]
\centerline{\includegraphics[width = \linewidth]{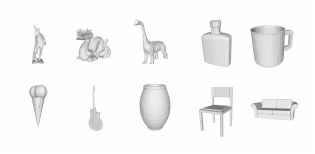}}
\caption{The ten novel objects used in the experiments. From left to right, top to bottom, the objects are neptune, dragon, noisy dino, bottle 2, mug 2, ice cream 2, guitar 2, vase 2, home chair and sofa 2 respectively.}
\label{fig:novel}
\end{figure}
Ten objects from the Princeton Shape Benchmark \cite{shilanePrincetonShapeBenchmark2004} and Stanford 3D Scanning Repository \cite{Stanford3DScanning} are selected as known objects, while ten comparable but distinct models from the same datasets are selected as novel objects (Figs.~\ref{fig:known}–\ref{fig:novel}). All models are scaled to approximately the same length and fit within a $6\times6\times6$ bounding box to avoid trivial discrimination based on the size of the object.

\subsection{Experimental Procedure}

For each object, 10 active exploration trials are performed using the proposed GPIS-DHD exploration procedure. Objects are initialized with random poses and remain stationary during each trial. After a random initial contact, the framework iteratively updates the belief over class and pose, reconstructs the surface when the object is classified as novel, and actively explores the object surface until the termination criterion is satisfied.

To show the exploration efficiency of the GPIS-DHD exploration procedure and robustness of the system under partial information, identical experiments are conducted using the RRT-based exploration procedure. Additionally, 10 trials are performed on a novel home chair model after adding its reconstructed shape to the known set, demonstrating incremental learning capability.

In total, $10 \times 20 \times 2 + 10 = 410$ trials are conducted.

\subsection{Evaluation Metric}
{We report the classification accuracy and the pose estimation error for trails for known objects. The pose estimation error is measured by the TWD between the MAP shape $\mathbf{M}$ and the ground truth object shape $\mathbf{G}$. 

For the novel object set, the reconstructed shapes using the proposed method (PF-MAP-GPIS) are compared to the MAP from the PF (PF-MAP) and the reconstructed shape using Screened Poisson reconstruction method\cite{kazhdanScreenedPoissonSurface2013} given the same contact sequences. The reconstruction error is measured by $\text{d}_{TH}(\mathbf{S}^{*},\mathbf{G)}$ the TWD between the reconstructed GPIS $\mathbf{S}^{*}$ and the ground truth object shape $\mathbf{G}$.

The surface uncertainty for both known and novel objects is instead measured by the DHD between $\mathbf{M}$ and all contact points $\mathbf{X}_c$.

In addition, we report the number of time steps taken to complete the trial for the proposed GPIS-DHD and RRT based exploration procedure respectively to compare their exploration efficiency. The maximal number of particles used in the experiments on the known object set is reported as an indicator for the sampling efficiency of the proposed sampling scheme. 

\subsection{Parameter Settings}

All trials use the parameter values listed in Table~\ref{tab1}. Noise parameters $\sigma_d$ and $\sigma_n$ are tuned on the known-object set. Sensory noise $\sigma$ matches the simulator’s zero-level tolerance ($0.01$ units). The GP kernel scaling parameter $a$ is initialized at $a_0=1$ and optimized online via one-step Newton's method updates.

\begin{table}[htbp]
\caption{Parameter settings.}
\begin{center}
\begin{tabular}{|c|c|c|c|c|c|c|c|c|}
\textbf{parameter} &\textbf{$\sigma_d$}& \textbf{$\sigma_n$}& \textbf{$\sigma$} & $a_0$ & $\epsilon$ & $n_s$ & $\lambda$\\
\hline
\textbf{value} & 0.50 & 1.50 & $1.00\times10^{-4}$ & 1.00 & 0.60 & 30  &  0.97\\
\end{tabular}
\label{tab1}
\end{center}
\end{table}

\section{Experiment Results}
For clarity, the analysis of the results is reported separately for known test objects and novel test objects. An overview of the results for the experiments can be found in Fig.~\ref{fig:results_overview}, which will be referred to in the following sections.

\subsection{Object Recognition and Pose Estimation with Known Objects}

In terms of object recognition, the framework identified the correct object class with 100\% accuracy regardless of the choice of the exploration procedures.

Regarding pose estimation, the framework achieved pose estimation errors below the desired threshold of 0.6 in 100 out of 100 trials with the GPIS-DHD procedure, and 99 out of 100 trials with RRT-based procedure. The failed case with RRT-based procedure took place when no contact points were acquired on the handle of the mug resulting in ambiguity in its pose. Side by side comparison of both procedures under the same initial conditions is shown in Fig.~\ref{fig:mug_diff}.


\begin{figure*}[!t] 
    \centering
  \subfloat[\label{fig:pose_estimation_error}]{%
    \includegraphics[width=0.25\linewidth]{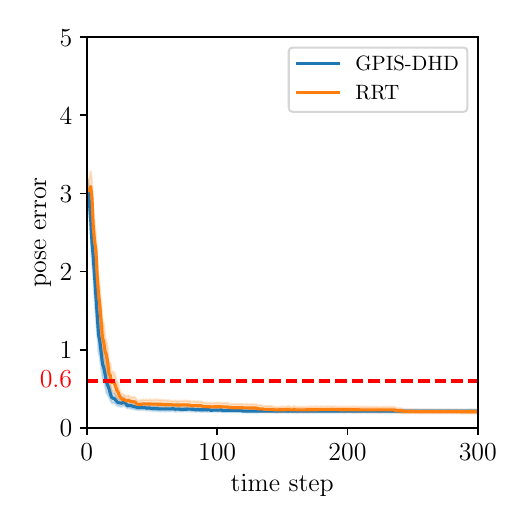}}
    \hfill
  \subfloat[\label{fig:desired_error_step}]{%
        \includegraphics[width=0.25\linewidth]{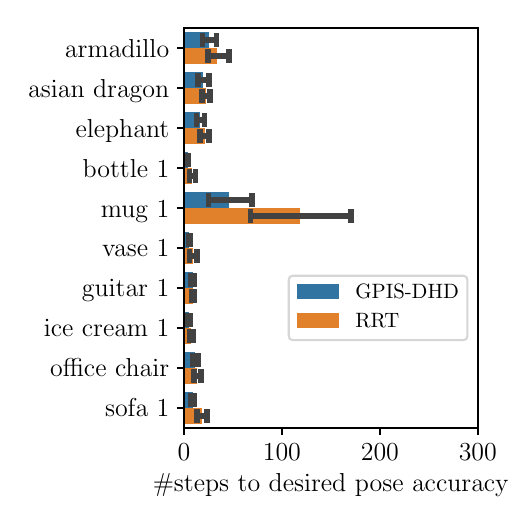}}
    \hfill
  \subfloat[\label{fig:directed_hausdorff_error}]{%
        \includegraphics[width=0.25\linewidth]{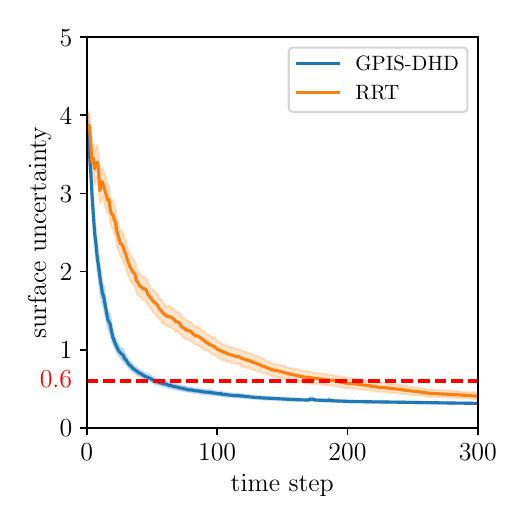}}
    \hfill
  \subfloat[\label{fig:termination_time_comparison}]{%
        \includegraphics[width=0.25\linewidth]{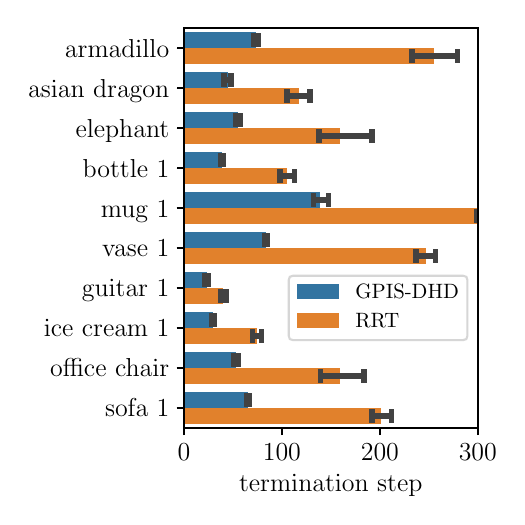}}
    \\
  \subfloat[\label{fig:average_reconstruction_error}]{%
        \includegraphics[width=0.25\linewidth]{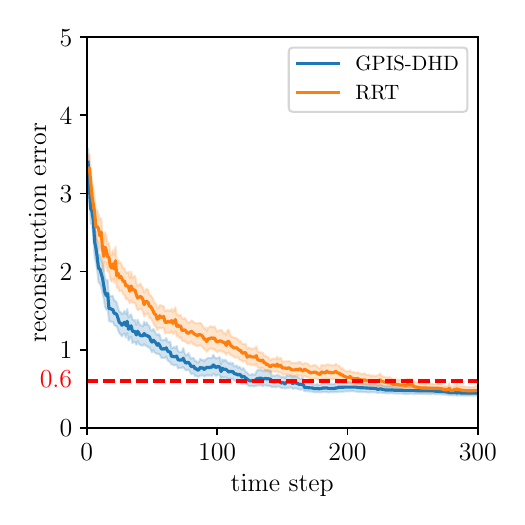}}
    \hfill
  \subfloat[\label{fig:sr_desired_error_step} ]{%
        \includegraphics[width=0.25\linewidth]{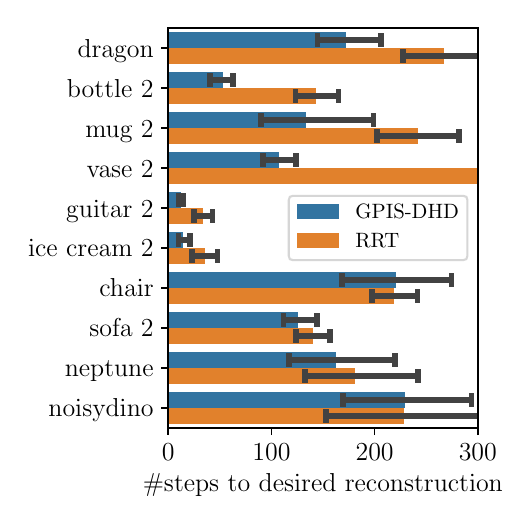}}
    \hfill
  \subfloat[\label{fig:sr_directed_hausdorff_error} ]{%
        \includegraphics[width=0.25\linewidth]{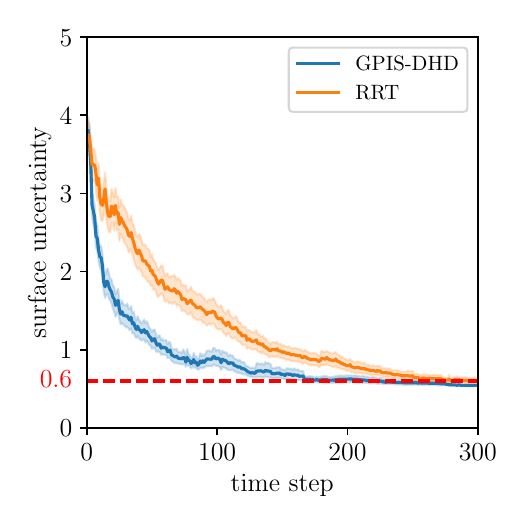}}
    \hfill
  \subfloat[\label{fig:sr_termination_time_comparison}]{%
        \includegraphics[width=0.25\linewidth]{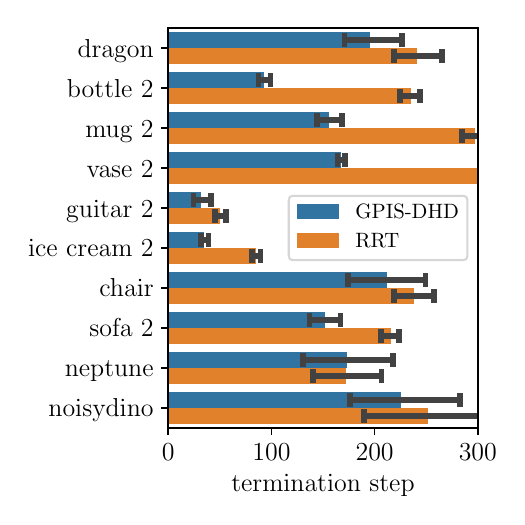}}
  \caption{An overview of simulation experiment results, the first row is for known objects and the second row is for novel objects: (a) Average pose estimation error over time with GPIS-DHD and RRT-based exploration procedures for known objects. The orange and blue line indicate the mean pose error at each time step over 100 trials for the ten known objects. The red dashed line indicates the desired pose error threshold. (b) The time steps to reach  below 0.6 pose error for each known object class. (c) Average DHD from MAP to contact data points over time with GPIS-DHD and RRT-based exploration procedures. The DHD from MAP shape to contact data points is used to determine if the contact data points cover the estimated surface sufficiently. The red dashed line indicates the DHD threshold for termination. (d) Termination time steps for known objects with GPIS-DHD and RRT-based exploration procedures. (e) Average shape reconstruction error for novel objects with GPIS-DHD and RRT-based exploration procedures for novel objects. The red dashed line indicates the desired reconstruction error threshold. (f) The time steps to reach below 0.6 reconstruction error for each novel object class. (g) Average DHD from MAP to contact data points over time with GPIS-DHD and RRT-based exploration procedures for novel objects. The DHD from GPIS to contact data points is used to determine if the contact data points cover the estimated surface sufficiently. The red dashed line indicates the DHD threshold for termination. (h) Termination time step for novel objects with GPIS-DHD and RRT-based exploration procedures.
  The translucent bands around the curves in (a) (c) (e) (g) and the error bars in (b) (d) (f) (h) indicate the confidence interval of 95\%. Despite the x axis in (b) (d) (f) (h) were clipped at 300 time steps for easier comparison, the experiments for known objects were stopped if they took beyond 300 time steps and the experiments for novel objects were stopped if they took beyond 400 time steps.}
  \label{fig:results_overview} 
\end{figure*}

As shown in Fig.~\ref{fig:pose_estimation_error}, the average pose estimation error falls below 0.6 within $\sim$20 steps for both exploration procedure. On the other hand, GPIS-DHD achieves faster surface coverage (Fig.~\ref{fig:directed_hausdorff_error}) and therefore satisfies the coverage-based termination criterion substantially earlier than RRT-based across all known objects (Fig.~\ref{fig:termination_time_comparison}). Class-wise, GPIS-DHD reaches the pose-error threshold earlier in 9 out of 10 classes, with the largest advantage for the mug (Fig.~\ref{fig:desired_error_step}), where locating the handle is required to resolve the remaining rotational symmetry (Fig.~\ref{fig:mug_diff}).

\begin{figure}[!t]
\centerline{\includegraphics[width = 0.8\linewidth]{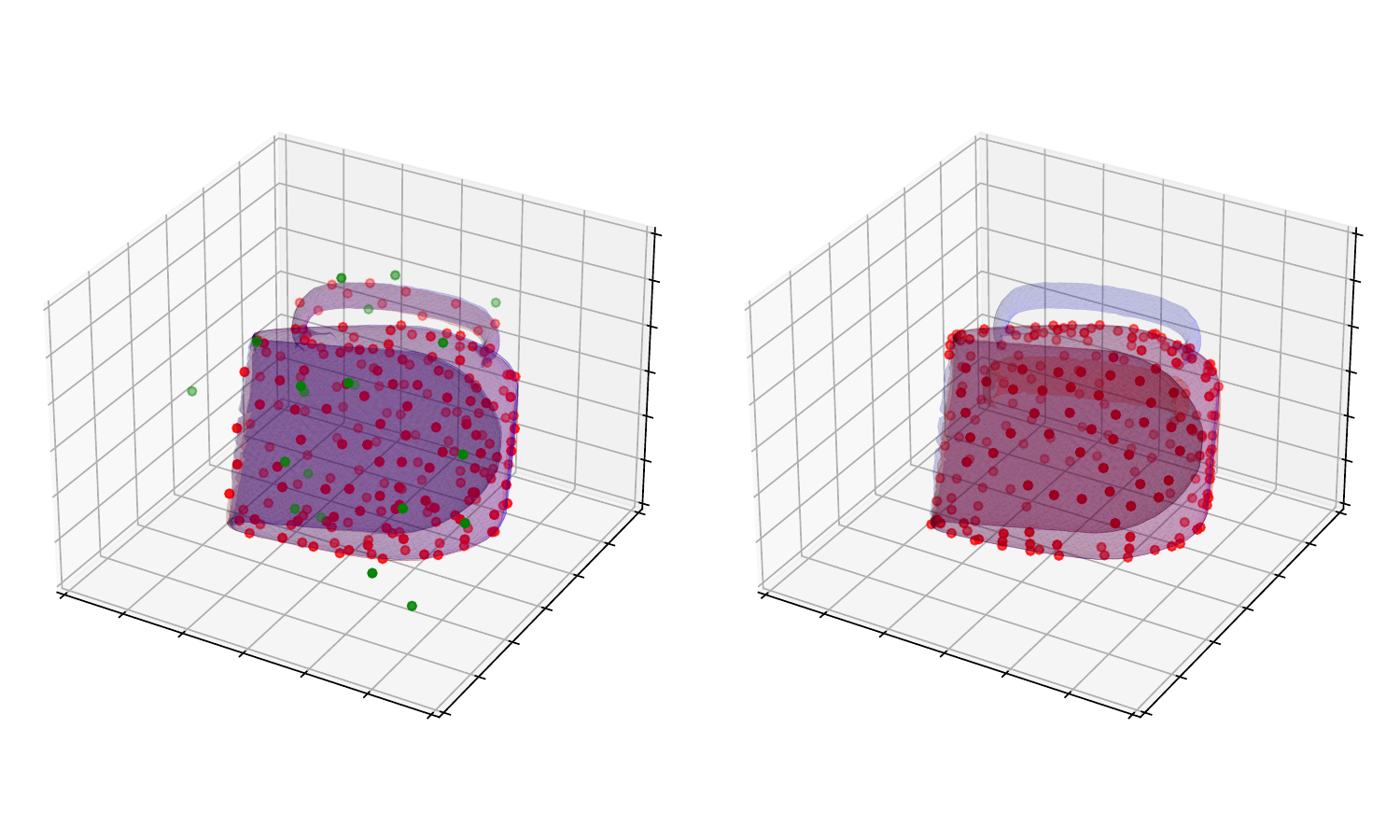}}
\caption{Comparison of the mug 1 case with GPIS-DHD and RRT exploration procedures respectively: On the left, GPIS-DHD exploration procedure managed to find the mug's handle after 200 time steps, therefore the ground truth (blue shape) and the MAP shape (red shape) overlapped. On the right, RRT exploration procedure failed to find the handle after 200 time steps, therefore lacked information to determine the mug's pose. Red points and green points are the contact points and non-contact points respectively.}
\label{fig:mug_diff}
\end{figure}

\subsection{Sampling Efficiency of the Particle Filter}
We tracked the number of particles during the GPIS-DHD object recognition and pose estimation experiments. The largest number of particles observed at any time step over 100 runs was 6914, representing the joint belief distribution of 10 object classes and the 6 DOF pose. This indicates that our customized particle filter kept the number of particles tractable at all time steps.

\subsection{Shape Reconstruction with Novel Objects}
\begin{table}[htbp]
\caption{Comparison of shape reconstruction methods}
    \centering
        \begin{tabular}{l|c|c|c|c|c|c}
        \hline
        {} & \multicolumn{6}{c}{Reconstruction error (TWD)} \\ 
        \hline
        {} & \multicolumn{2}{c|}{PF-MAP} & \multicolumn{2}{c|}{PF-MAP-GPIS} & \multicolumn{2}{c}{Poisson} \\
        \hline
        {class} &   mean &       std &     mean &       std &     mean &      std \\
        \hline
        bottle 2    &        0.994 &  0.088 &   \textbf{0.259} &  \textbf{0.036} &      0.416 &  0.063 \\
        dragon      &        2.952 &  0.133 &   \textbf{0.478} &  \textbf{0.057} &      0.582 &  0.122 \\
        guitar 2    &        \textbf{0.412} &  \textbf{0.062} &   \textbf{0.412} &  \textbf{0.062} &      1.092 &  0.673 \\
        home chair  &        1.771 &  \textbf{0.085} &   \textbf{0.505} &  0.402 &      1.054 &  0.360 \\
        ice cream 2 &    \textbf{0.432} &  \textbf{0.046} &    \textbf{0.432} &  \textbf{0.046} &      0.546 &  0.126 \\
        mug 2       &        0.580 &  \textbf{0.129} &   \textbf{0.526} &  0.335 &      1.703 &  0.282 \\
        neptune     &        1.735 &  0.801 &   \textbf{0.505} &  \textbf{0.125} &      0.655 &  0.201 \\
        noisydino   &        1.053 &  0.106 &   0.519 &  \textbf{0.089} &      \textbf{0.518} &  0.152 \\
        sofa 2      &        0.611 &  0.049 &   \textbf{0.483} &  \textbf{0.043} &      0.604 &  0.045 \\
        vase 2      &        1.453 &  0.221 &   \textbf{0.335} &  \textbf{0.060} &      0.658 &  0.081 \\
        \hline
\end{tabular}

\label{tab:sr_comparison}
\end{table}

\begin{figure}[!t] 
\centering
    \includegraphics[width=\linewidth]{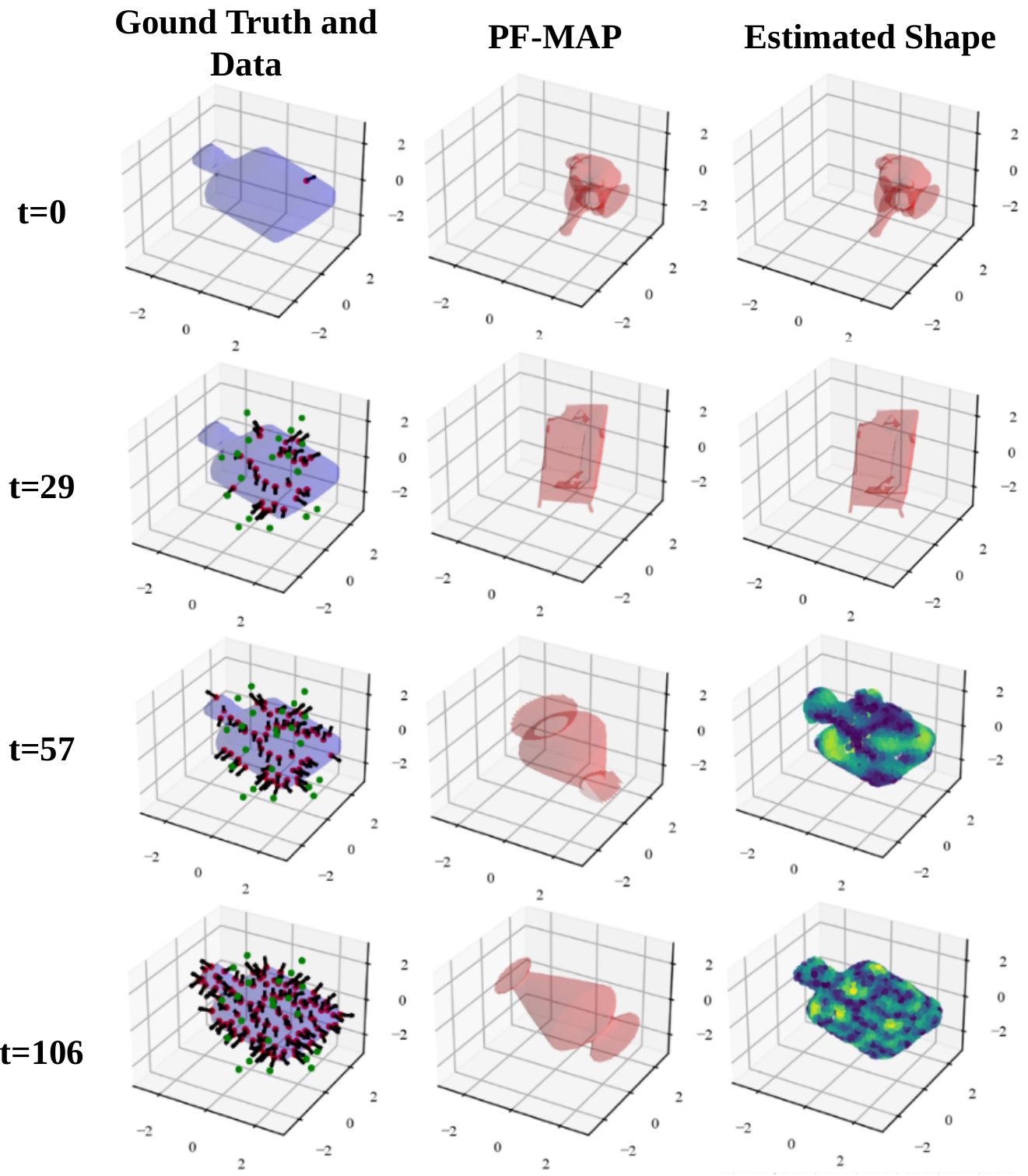}
    \hfill
  \caption{An example of the inference process of the proposed framework over time. At the beginning of the inference process ($t=0$ and $t=29$), PF-MAP fits the data well and the object is regarded as known, the estimated shape is the same as PF-MAP. After more data are collected ($t=57$), the framework identifies the object as a novel object and reconstruct the shape with PF-MAP-GPIS. It continues to refine the prior and shape with PF-MAP-GPIS till the desired data coverage on the estimated shape is achieved ($t=106$). For $t=57$ and $t=106$, brighter color on the estimated surface indicates larger GPIS posterior variance (uncertainty).}
  \label{fig:inference_process}
\end{figure}

Table \ref{tab:sr_comparison} summarizes the statistics of the reconstruction error for each class in the novel object set. Each column reports the statistics of the TWD between the reconstructed shape with the corresponding method and the ground truth shape. PF-MAP denotes using the MAP from the PF (the prior for our proposed method), PF-MAP-GPIS denotes our method, and Poisson denotes the Screened Poisson reconstruction method (depth = $9$).  The same oriented point clouds acquired from the GPIS-DHD exploration experiments are used for all reconstruction methods. In most cases, PF-MAP-GPIS yield lower reconstruction error than Poisson and substantially improves upon PF-MAP. Notably, two novel objects, ice cream 2 and guitar 2, are classified as known objects, as the geometric differences between them and their known object variants (ice cream 1 and guitar 1) are not large enough. 
In these two cases, PF-MAP-GPIS returns the same results as PF-MAP.

\begin{figure}[!t]
\centerline{\includegraphics[width = 0.8\linewidth]{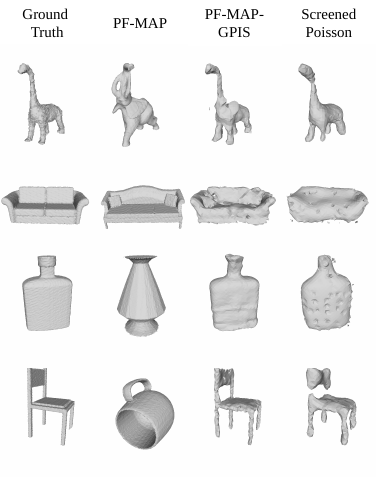}}
\caption{Examples of the maximum a posteriori (MAP) priors and reconstructed surfaces in the experiments. Each row is one experiment. From left to right, the first column shows the ground truth object shape, the second column shows the MAP shape from the particle filter (PF-MAP), the third column shows the Gaussian process implicit surface reconstructed using the MAP shape as prior (PF-MAP-GPIS), the last column shows the reconstructed result of Screened Poisson surface reconstruction (Screened Poisson).}
\label{fig:reconstruction_priors}
\end{figure}
 Fig~\ref{fig:inference_process} shows the inference process during an experiment on the novel object set, where the framework first identifies the novel object, then gradually refines the shape prior used for GPIS reconstruction and the reconstructed shape as more data are collected.
Representative examples of GPIS shape reconstruction with MAP shape priors are shown in Fig~\ref{fig:reconstruction_priors}, where it can be seen the proposed framework makes sensible choices of the MAP priors and the GPIS reduces the discrepancies between the MAP priors and the ground truth shapes.




Compared to pose estimation for known objects, achieving the desired reconstruction accuracy for novel objects requires more time steps (Figs.~\ref{fig:desired_error_step} and \ref{fig:sr_desired_error_step}). On average, GPIS-DHD reaches the desired reconstruction error threshold faster than RRT (Fig.~\ref{fig:average_reconstruction_error}), and achieves earlier surface coverage (Figs.~\ref{fig:sr_directed_hausdorff_error}) and termination in most classes (Fig.~\ref{fig:sr_termination_time_comparison}).
\subsection{Including Learned Shapes as Priors}
To demonstrate the incremental learning capability of the framework, we included the learned home chair model as a new prior (in total 11 known objects to the framework) and then performed experiments on the home chair in different poses with the GPIS-DHD exploration procedure. In total, 10 trials were carried out. The results are shown in Figs.~\ref{fig:new_prior}, \ref{fig:PE_learned} and \ref{fig:DHD_learned}. In 10 out of 10 trials (100\%), the framework successfully recognized the chair and reached the desired pose estimation error within 50 time steps. On average, it took approximately $68$ steps to finish the exploration, which is significantly faster compared to more than 200 steps before adding the learned chair prior (Fig.~\ref{fig:sr_termination_time_comparison}). 
\begin{figure}[!t]
\centerline{\includegraphics[width = 0.8\linewidth]{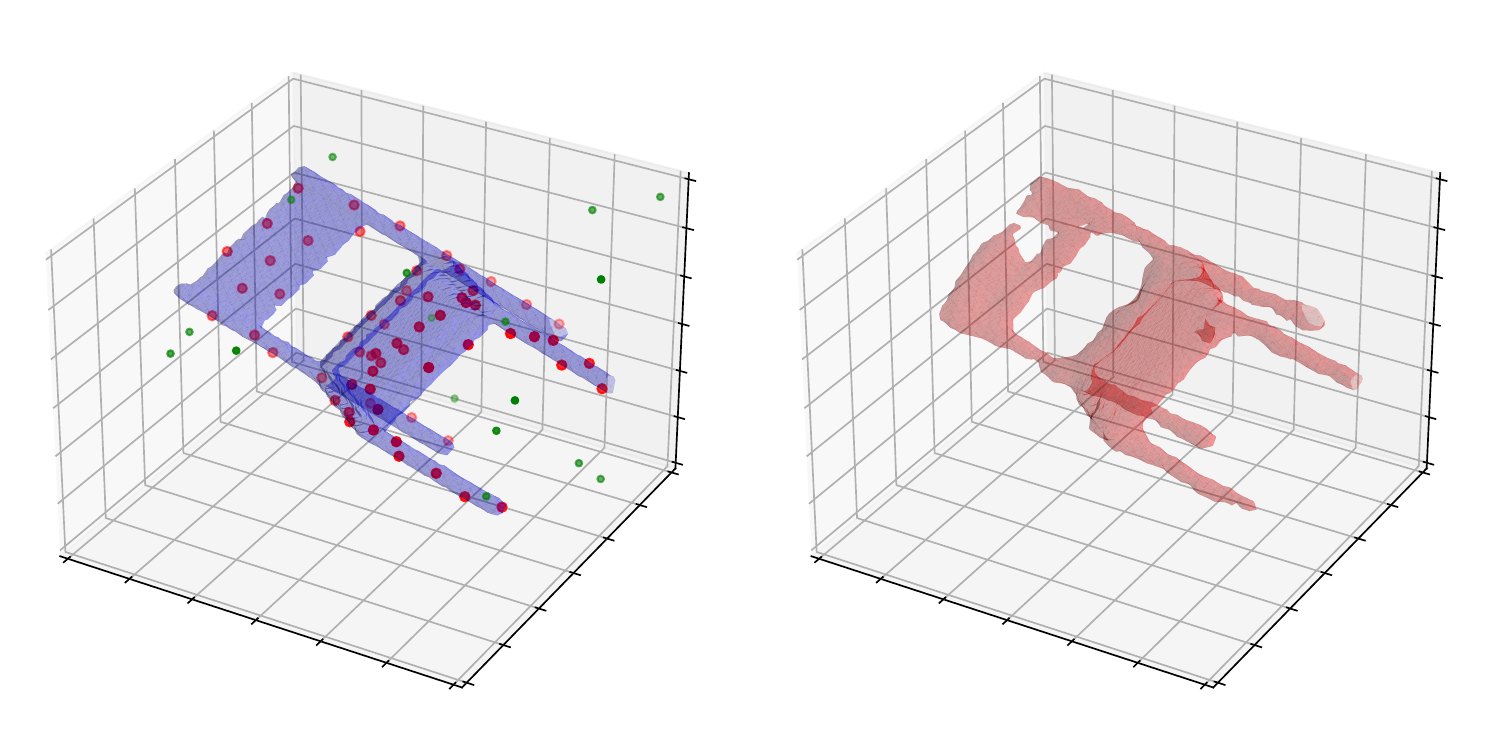}}
\caption{An example of including a learned shape as a prior of the framework: After including the learned chair shape as a prior, the framework successfully recognized the chair (blue shape) as the learned reconstructed model (red shape).}
\label{fig:new_prior}
\end{figure}
\begin{figure}[!t] 
\centering
  \subfloat[\label{fig:PE_learned}]{%
        \includegraphics[width=0.50\linewidth]{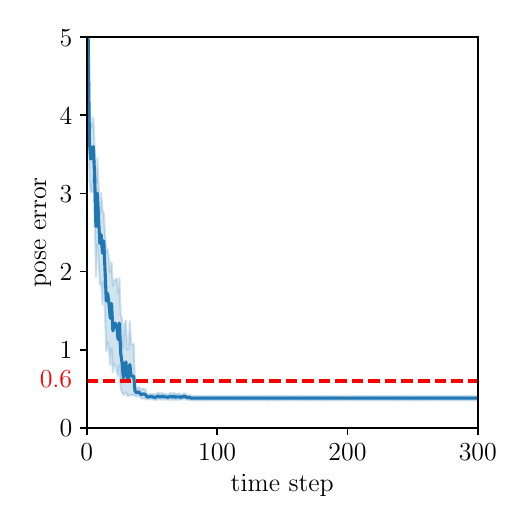}}
    \hfill
  \subfloat[\label{fig:DHD_learned}]{%
        \includegraphics[width=0.50\linewidth]{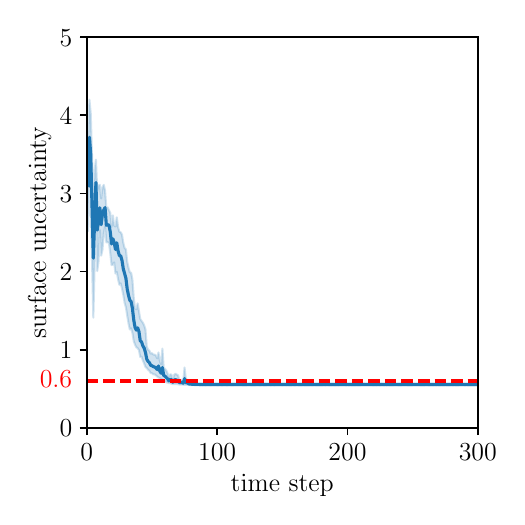}}
  \caption{(a) Average pose estimation error over 10 trials in the learned chair prior experiment. The red dashed line indicates the pose error threshold. (b) Average DHD error 10 trials in the learned chair prior experiment. The red dashed line indicates the DHD threshold for termination. The translucent bands around the curves in (a) (b) indicate the confidence interval of 95\%.}
  \label{fig2} 
\end{figure}
The results show that the framework can learn new object shapes through shape reconstruction and utilize the knowledge effectively for object recognition and pose estimation, even though the learned shape is not identical to the actual object.

\section{Discussion}
In this work, we presented a Bayesian framework that efficiently identifies object class and pose while transferring knowledge from known shapes to learn novel shapes through active touch. Newly learned shapes extend the prior, enabling the framework to successfully recognize objects in subsequent trials. We frame object recognition, shape, and pose estimation as a non-linear Bayesian filtering problem and employ a particle filter-based approach (PF).

\subsection{Object Recognition and Pose Estimation for Known Objects}
Compared to Kalman Filters, PFs can approximate arbitrary distributions, which is needed to capture the ambiguity and uncertainty present due to partial tactile observations. Our PF uses a progressive sampling strategy based on point-pair features and a custom weight assigning scheme to achieve efficient estimations of the joint distribution of object class and object pose. 
The core idea behind the point-pair-feature-based sampling procedure is to use translation and rotation invariant features, in our case, distances and angles, to find possible point correspondences, and then use the point correspondences to solve for the potential object poses, making it more efficient than considering single points. This idea is not limited to the point-pair feature. For instance, three or four points could be considered simultaneously. Nonetheless, a set of two points is chosen in this study because the alignment of two non-bilinear oriented point pairs can uniquely determine the pose while the number of all point pair combinations is tractable. If more than two points per set are considered simultaneously, the number of possible combinations quickly becomes intractable. Similarly, in the sampling scheme, each object was represented by 200 points to keep the number of combinations tractable. Increasing the number of points will likely make the framework achieve the desired pose accuracy sooner but at the cost of higher computational cost.

In the object recognition and pose estimation experiment with known objects, the framework recognized the correct object class and achieved the desired pose estimation error with a small number of data points, regardless of the selected exploration procedure. Both the GPIS-DHD and RRT-based exploration procedures worked well, however GPIS-DHD generally covered the estimated surface faster than the RRT-based exploration procedure. The mug example highlights the benefit of the GPIS-DHD exploration procedure: prioritizing global data coverage of the estimated surface and incorporating non-contact observations helps resolve symmetry ambiguities by actively locating the asymmetric part of the object, in this case, the handle of the mug. The DHD-based termination criterion further ensured that exploration continued until such ambiguities were resolved, in this case, contact points were collected on the handle. 

\subsection{Shape Reconstruction for Novel Objects}
The PF can identify novel objects but lacks the ability to learn new shapes, so the GPIS is introduced to perform shape reconstruction. Aiming to exploit prior knowledge, we initiate the prior of the GPIS with the MAP shape from the PF. Our results showed that GPIS with MAP priors worked effectively with the novel object set, and it achieved better performance than the screened Poisson surface reconstruction with the sparse point clouds from the GPIS-DHD exploration procedure, indicating the importance of prior knowledge in low-data regimes.

Interestingly, GPIS achieved good reconstructions even when the MAP prior differed substantially from the ground truth (see Table~\ref{tab:sr_comparison}). This suggests that local geometric similarities between the prior and the actual shape can provide valuable information for efficient shape learning, with GPIS preserving similar regions while correcting inconsistent regions based on observed data.

 In general, the framework took longer to explore a novel object than a known object.  Active tactile exploration for ice cream 2 and guitar 2 was merely slightly slower than the known ice cream 1 and guitar 1 with small geometric differences between priors and the actual objects, while for other novel objects, it was substantially slower. That is in part due to the varying applicability of the prior knowledge for these novel objects. 

In this study, the framework determined if an object was known or novel based on the MAP model evidence from the PF. If the object does not deviate enough from all known objects, it would be recognized as a known object. As was the case for ice cream 2 and guitar 2 in our set. This threshold for the decision can vary to provide some flexibility in choosing how similar the object should be to be classified as a known object. A different threshold could be used depending on the level of difference on the shape permitted by the task. The threshold used in this study should be interpreted as a heuristic. 



\subsection{Termination Criterion}
Although DHD was used as the termination criterion in this study, it should be viewed more generally as a measure of surface coverage. Lower DHD thresholds increase exploration time but improve coverage, while higher DHD thresholds allow faster decisions at the cost of potential errors. Importantly, DHD provides an uncertainty-aware estimate of surface exploration completeness, which remains informative even if alternative stopping criteria e.g., time limit and number of actions, are used.

\subsection{Computational Cost}
When it comes to computational complexity, our PF's computational cost does not scale linearly with the number of observations, because the PF's weight assignment scheme for newly sampled particles only requires the evaluation of two particles (current MAP particle and new MAP particle) per class on all previous observations, while for other new particles, it is a fixed cost. However, as more learned shapes are added to the prior set, it will require proportionally more memory and computation time.

The GPIS also introduces a computational bottleneck, with cubic complexity in the number of observations. 
Since the prior of the GPIS can be different at each time step, sparse GP with inducing points\cite{snelsonSparseGaussianProcesses2005,titsiasVariationalLearningInducing2009} is not directly applicable. A potential solution is to fix the prior after sufficient data coverage on the estimated surface, enabling sparse approximations.
The disadvantage of the increasing computational cost over time was partially compensated by using the proposed exploration procedure and the termination criterion, since the exploration terminated before the number of points would became untractable.



\subsection{Limitations and Future Work}
Several limitations remain in the current framework. 
Firstly, GPIS reconstruction occasionally produce defects such as small holes on thin surfaces and disconnected small parts, which is reflected in the reported reconstruction error. Preliminary experiments with alternative kernels (e.g., RBF) did not outperform the thin-plate kernel, although a systematic kernel ablation remains to be investigated. Addressing these undesired defects in the learned models will be valuable future work.

Secondly, the current framework assumes a single static object and does not consider object motion or multi-object scenes. Extending the Bayesian formulation to dynamic environments and multiple interacting objects is a natural next step.

Thirdly, both memory usage and computational cost of the PF scale linearly with the number of priors, which may limit scalability as more shapes are learned. Developing more compact or hierarchical representations of priors would improve scalability to larger prior sets.

Lastly, all experiments were conducted in a simplified simulation environment to validate the proposed framework. Ongoing work aims to transfer the approach to a real robotic platform, which requires integration with a complete hybrid control and motion planning framework and non-trivial realistic adaptions of the current framework, therefore lies beyond the scope of this paper. Although this study focuses on active tactile exploration, the framework is naturally extensible to multimodal perception, and future work will investigate joint visuo-tactile integration within the same Bayesian framework.

\section{Conclusion}

This paper presented a unified Bayesian framework for active tactile object recognition, pose estimation, and shape transfer learning. By combining a particle-filter-based formulation for joint object class and pose inference with GPIS-based shape reconstruction, the proposed approach enables a single probabilistic framework to address both object identification, localization, and novel shape learning through active tactile exploration.

Simulation experiments demonstrated that the framework reliably recognizes known objects and estimates their poses using sparse tactile observations, while transferring prior shape knowledge to reconstruct novel objects. Furthermore, newly reconstructed shapes can be incorporated as priors, enabling incremental learning and enhancing active exploration efficiency.

Overall, the proposed framework provides a principled probabilistic approach toward uncertainty-aware robotic perception that integrates recognition, localization, and shape learning within a single active sensing loop. We believe that the proposed framework is a step toward a general and robust robotic perception system capable of efficient continuous learning utilizing prior knowledge.

\section*{Acknowledgement}
We would like to express our gratitude to Dr. Lorenzo Natale for his valuable feedback on the paper.

\bibliographystyle{IEEEtran} 
\bibliography{IEEEabrv,citing_papers}
\vskip 0pt plus -1fil
 \begin{IEEEbiography}[{\includegraphics[width=1in,height=1.25in,clip,keepaspectratio]{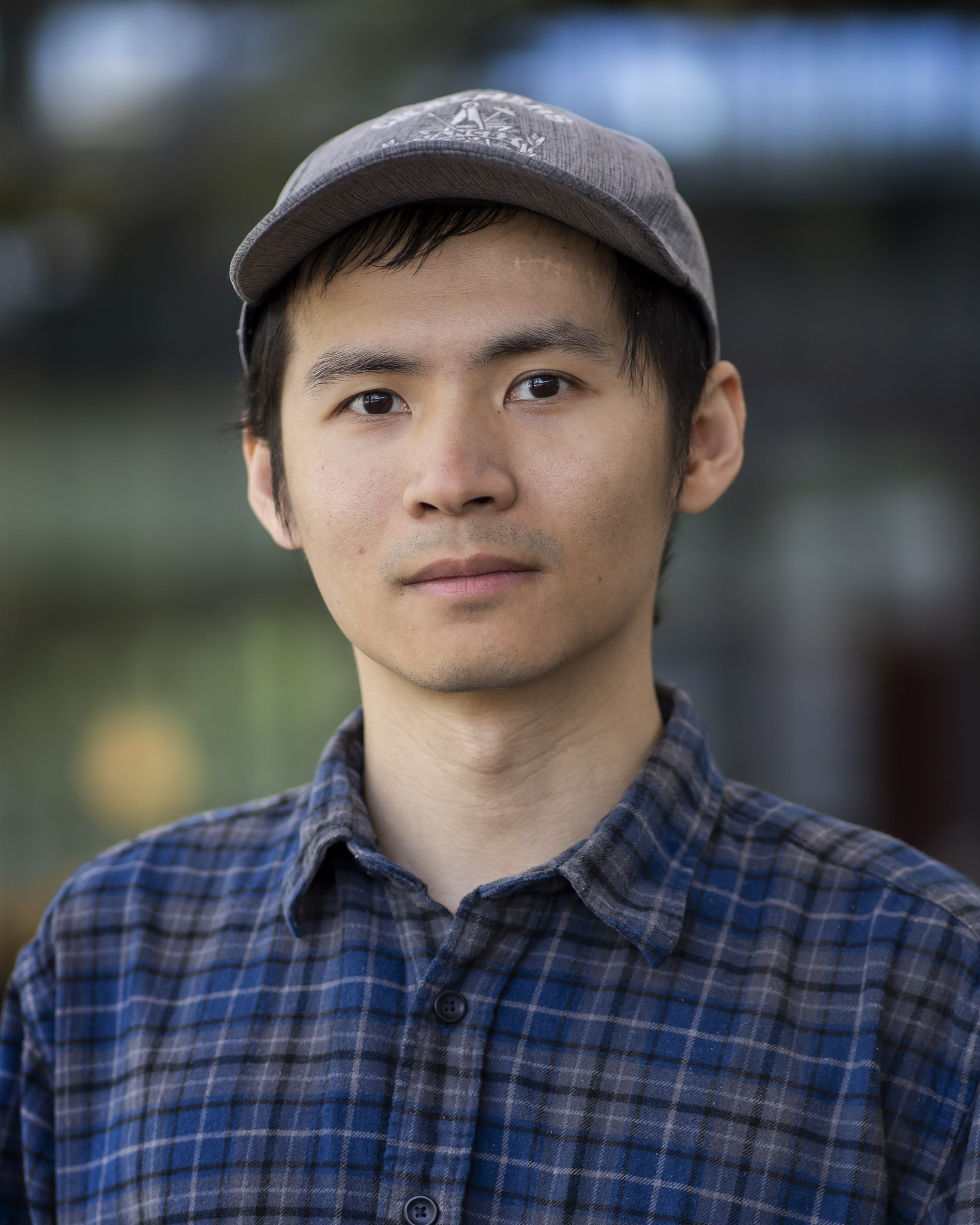}}]{Haodong Zheng} received his B.Eng. Degree in Mechatronic Engineering in 2017 at the Shantou University, China, and his M.Sc. Degree in Complex Adaptive Systems in 2022 at the Chalmers University of Technology, Sweden. He is currently a PhD candidate in the Human Technology Interaction Group at the Eindhoven University of Technology, The Netherlands. His research interests include robotic visuo-tactile perception and manipulation through Deep Learning and Bayesian methods. 
\end{IEEEbiography}
\vskip 0pt plus -1fil

\begin{IEEEbiography}[{\includegraphics[width=1in,height=1.25in,clip,keepaspectratio]{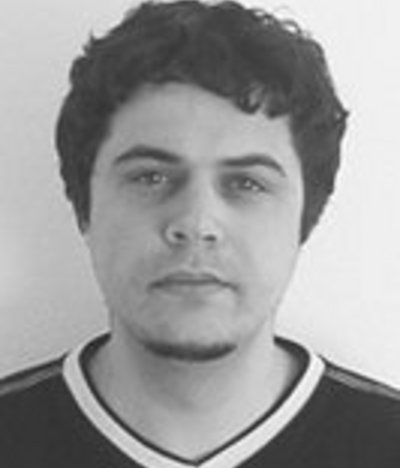}}]
{Andrei C. Jalba} received his Ph.D. degree in 2004 from the Institute for
Mathematics and Computing Science of the University of Groningen, The
Netherlands. Currently, he is an assistant professor in visualization and computer graphics at the Eindhoven University of Technology, the Netherlands. 
His expertise includes multi-scale shape representation, processing 
and reconstruction, physics-based material simulation and visual,
scientific and multi-core computing.
\end{IEEEbiography}

\vskip 0pt plus -1fil
\begin{IEEEbiography}
[{\includegraphics[width=1in,height=1.25in,clip,keepaspectratio]{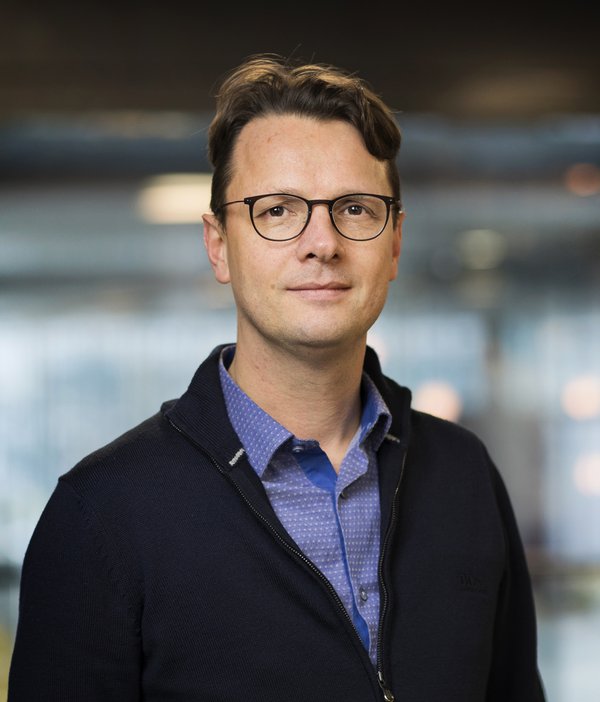}}]
{Raymond H. Cuijpers} graduated in Applied Physics at the TU/e in 1996. He received his PhD in Physics of Man from Utrecht University in 2000. He did a postdoc on the role of shape perception on human visuo-motor control at Erasmus MC Rotterdam. In 2004, he did a second postdoc at Radboud University Nijmegen. In 2008 he was appointed Assistant Professor at TU/e at the Human-Technology Interaction group. In 2014 he became associate professor in Cognitive Robotics and Human-Robot Interaction. His research interests include visual perception and visuo-motor control.
\end{IEEEbiography}

\vskip 0pt plus -1fil
\begin{IEEEbiography}[{\includegraphics[width=1in,height=1.25in,clip,keepaspectratio]{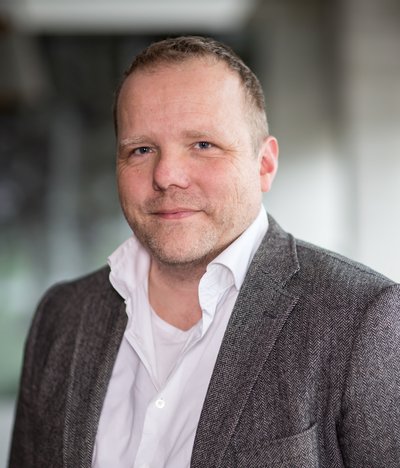}}]
{Wijnand IJsselsteijn} is a full professor of Cognition and Affect in Human-Technology Interaction at Eindhoven University of Technology (TU/e). He has an active research program on the impact of media technology on human psychology, and the use of psychology to improve technology design. His focus is on conceptualizing and measuring human experiences in relation to digital environments (immersive media, serious games, affective computing, personal informatics) in the service of human learning, health, and wellbeing. He has a keen interest in the relation between data science, AI and psychology, and works on technological innovations (such as sensor-enabled mobile technologies, virtual environments) that make possible novel forms of human behavior tracking, combining methodological rigor with ecological validity.
\end{IEEEbiography}
\vskip 0pt plus -1fil

\begin{IEEEbiography}[{\includegraphics[width=1in,height=1.25in,clip,keepaspectratio]{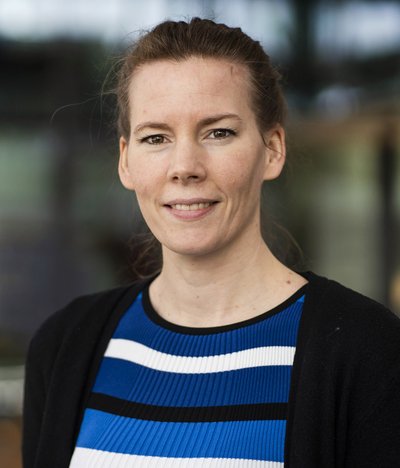}}] 
{Sanne Schoenmakers} received her Ph.D. degree in 2019 from the Artificial Intelligence department and Donders Institute for brain cognition and behaviour from the Radboud University in Nijmegen, the Netherlands. Currently, she is an assistant professor at Human Technology Interaction on Artificial Intelligence and Human Computer Interaction at the Eindhoven University of Technology, the Netherlands. Her expertise includes Bayesian modelling and Bayesian inference, computational cognitive neuroscience, and artificial intelligence.
\end{IEEEbiography}
\end{document}